\pdfoutput=1

\documentclass[11pt]{article}

\usepackage[preprint]{acl}

\usepackage{times}
\usepackage{latexsym}

\usepackage[T1]{fontenc}

\usepackage[utf8]{inputenc}

\usepackage{microtype}

\usepackage{inconsolata}

\usepackage{hyperref}
\usepackage{url}

\usepackage{amsmath,amsfonts}
\usepackage{algorithmic}
\usepackage{algorithm}
\usepackage{array}
\usepackage{stfloats}
\usepackage{textcomp}
\usepackage{verbatim}
\usepackage{amsthm}
\usepackage{anyfontsize}
\usepackage{float}            
\usepackage{mathrsfs}         
\usepackage{bm}              
\usepackage{threeparttable}   
\usepackage{makecell}      

\usepackage{booktabs}
\usepackage{multirow}
\usepackage{wrapfig}
\usepackage{balance}
\usepackage{colortbl}

\usepackage{graphicx}
\usepackage{multicol} 
\usepackage{subcaption}
\usepackage{enumitem}
\usepackage{float}
\usepackage{xcolor}
\usepackage[markup=default,authormarkup=none,commandnameprefix=always]{changes}
\usepackage{longtable} 
\usepackage{placeins}

\newcommand{\sysname}[0]{\textcolor{black}{\textsc{Conf-Profile}}}

\title{\textsc{Conf-Profile}: A Confidence-Driven Reasoning Paradigm \\for Label-Free User Profiling}

\author{
 \textbf{Yingxin Li\textsuperscript{1,2}\thanks{Equal contribution.}\thanks{Work during internship at Bytedance.}},
 \textbf{Jianbo Zhao\textsuperscript{1}}\footnotemark[1], 
 \textbf{Xueyu Ren\textsuperscript{1}},
 \textbf{Jie Tang\textsuperscript{1}},
 \textbf{Wangjie You\textsuperscript{1}},
 \textbf{Xu Chen\textsuperscript{1}},
  \\
 \textbf{Kan Zhou\textsuperscript{1}},
 \textbf{Chao Feng\textsuperscript{1}}\thanks{Corresponding to: 
   \href{mailto:chaofeng.zz@bytedance.com}{chaofeng.zz@bytedance.com}\\
   \href{mailto:wangzhi@sz.tsinghua.edu.cn}{wangzhi@sz.tsinghua.edu.cn}},
 \textbf{Jiao Ran\textsuperscript{1}},
 \textbf{Yuan Meng\textsuperscript{3}},
 \textbf{Zhi Wang\textsuperscript{2}}\footnotemark[3]
\\
 \textsuperscript{1} Douyin Content Group, Bytedance \\
 \textsuperscript{2} Shenzhen International Graduate School, Tsinghua University \\
 \textsuperscript{3} Department of Computer Science and Technology, Tsinghua University 
}

\begin{document}
\maketitle
\begin{abstract}

User profiling, as a core technique for user understanding, aims to infer structural attributes from user information.
Large Language Models (LLMs) provide a promising avenue for user profiling, yet the progress is hindered by the lack of comprehensive benchmarks.
To bridge this gap, we propose ProfileBench, an industrial benchmark derived from a real-world video platform, encompassing heterogeneous user data and a well-structured profiling taxonomy.
However, the profiling task remains challenging due to the difficulty of collecting large-scale ground-truth labels, and the heterogeneous and noisy user information can compromise the reliability of LLMs.
To approach label-free and reliable user profiling, we propose a \underline{Conf}idence-driven \underline{Profile} reasoning framework \textbf{\sysname}, featuring a two-stage paradigm. 
We first synthesize high-quality labels by leveraging advanced LLMs with confidence hints, followed by confidence-weighted voting for accuracy improvement and confidence calibration for a balanced distribution.
The multiple profile results, rationales, and confidence scores are aggregated and distilled into a lightweight LLM.
We further enhance the reasoning ability via confidence-guided unsupervised reinforcement learning, which exploits confidence for difficulty filtering, quasi-ground truth voting, and reward weighting.
Experimental results demonstrate that \sysname~ delivers substantial performance through the two-stage training, improving F1 by 13.97 on Qwen3-8B.
\end{abstract}

\section{Introduction}
\label{section:intro}

\setlist{topsep=0pt, partopsep=0pt, parsep=0pt, itemsep=0.1pt}

\begin{figure}[t]
    \centering
    \includegraphics[width=\columnwidth]{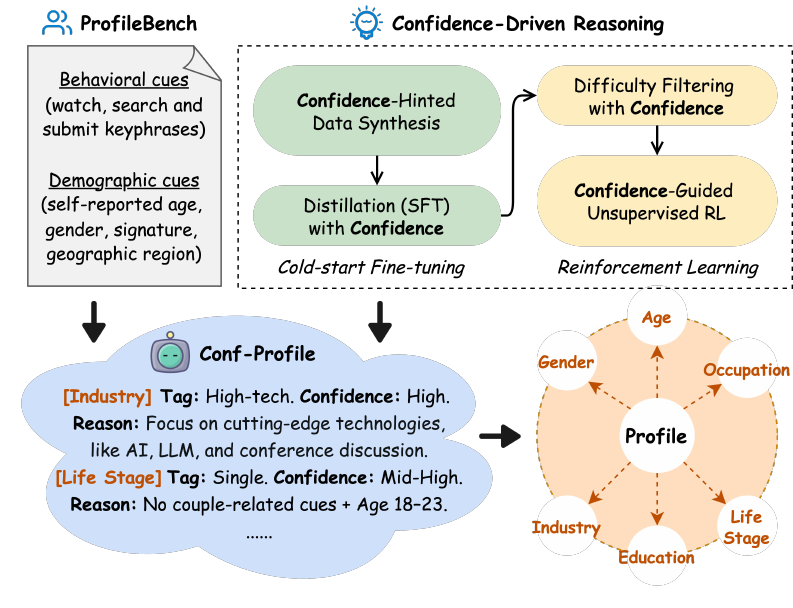}
    \caption{
    \textbf{Overview of \sysname.} We introduce ProfileBench, a novel benchmark consisting of heterogeneous cues for the user profiling task. Through a two-stage confidence-driven reasoning framework, we develop \sysname, which models six core attributes.
    }
    \label{fig: intro_workflow}
\end{figure}

User profiling seeks to infer structural attributes of users, serving as a fundamental step for user understanding and personalized services \cite{user-understand, user-understand-survey, user-personalization}.
The vast amount of user-generated content on the Internet offers rich evidence to construct virtual user representations and reveal underlying preferences. 
User profiles along dimensions including age, gender, industry, occupation, education level and life stage are critical to personalization, powering social network analysis \cite{profile-for-social, profile-for-social-dl}, recommendation systems \cite{profile-for-rec, profile-for-chat-rec}, misinformation mitigation \cite{profile-for-fake}, and large-scale surveys \cite{profile-for-survey}. 
However, existing approaches remain limited.
Survey-based methods are costly and often incomplete, while static registration data quickly becomes outdated \cite{profile-with-survey}. 
Supervised learning pipelines struggle with the scarcity of labeled data, and most methods narrowly focus on a single attribute, resulting in fragmented and unreliable user understanding \cite{profile-survey-sup}. 

Large Language Models (LLMs) \cite{openai2024gpt4technicalreport, deepseekai2025deepseekr1incentivizingreasoningcapability, yang2025qwen3technicalreport} provide a promising alternative, as they can handle diverse inputs and jointly infer multi-dimensional profiling tags.
Recent studies have explored this potential by either inferring profile labels via prompting and fine-tuning \cite{profile-construct-update-benchmark, profile-youarewhatyoubought, profile-tuning-emb}, or by synthesizing user data to construct virtual profiles \cite{profile-construct-update-benchmark, profile-persona}. 
Despite these advances, their application to user profiling faces several fundamental challenges.
First, existing works are dominated by synthetic datasets, which fail to capture the real-world complexity and leave the field without a credible, standardized benchmark.
Second, collecting large-scale ground-truth profiles is prohibitively expensive and often infeasible, underscoring the necessity of developing label-free methods.
Third, precise profiling requires reasoning over heterogeneous and noisy signals, yet current prompt-based or distillation approaches largely rely on shallow pattern matching, compromising reliability and inducing unwarranted certainty. 
Collectively, these challenges motivate the need for a framework that not only reasoning without labels but also explicitly manages uncertainty.

To address these issues, we propose \textbf{\sysname}, a confidence-driven framework for label-free user profiling, together with ProfileBench, a realistic benchmark for standardized evaluation.
\emph{ProfileBench} is an industry-derived dataset from a large-scale video platform, \textit{Douyin}. It integrates heterogeneous user information, defines a systematic profiling taxonomy, and provides 1,000 curated human-annotated samples, with all data authorized and anonymized for privacy compliance.
Building upon this benchmark, \sysname~ employs confidence as an uncertainty signal throughout the pipeline. 
Specifically, it begins with a \emph{confidence-hinted data synthesis pipeline} to produce high-quality pseudo-labels. 
This pipeline performs parallel sampling to gather diverse candidate labels and confidence scores, applies confidence-weighted voting to aggregate robust tags, and calibrates the confidence score for profiling task adaptation.
The resulting voted tags, calibrated confidence, and rationales are distilled into a lightweight LLM, enabling efficient training and deployment. 
To further enhance the reasoning ability, \emph{confidence-guided unsupervised reinforcement learning} is applied, where difficulty filtering highlights informative samples, multi-rollout voting with confidence produces quasi-ground truth, and calibrated confidence serves as adaptive reward weights. 
Together, these designs improve profiling robustness while ensuring reliable confidence estimates.

We conduct experiments to validate the effectiveness of \sysname~  on ProfileBench. 
Experimental results show that by applying the two-stage confidence-driven training paradigm on Qwen3-8B, \sysname~ achieves consistent F1 gains of 10.61 and 3.36, outperforming state-of-the-art (SOTA) reasoning LLMs.
Our contributions are summarized as follows:
\begin{itemize}
\item [(1)] \textbf{ProfileBench benchmark.} We introduce the first industry-level benchmark for user profiling, integrating heterogeneous user cues with a systematic six-dimensional taxonomy, enabling standardized evaluation.
\item [(2)] \textbf{Confidence-hinted data synthesis.} We design a pipeline that generates pseudo-labels with rationales and calibrated confidence scores, alleviating label scarcity and providing reliable cold-start supervision.
\item [(3)] \textbf{Confidence-guided unsupervised reinforcement learning.} We propose an optimization scheme that leverages confidence to filter samples, vote for targets, and reshape rewards, strengthening the reasoning ability.
\item [(4)] \textbf{Comprehensive validation.} We conduct extensive experiments, which demonstrate consistent gains on ProfileBench and reveal the critical role of confidence, underscoring the broader impact of \sysname.
\end{itemize}

\section{Related Work}

\subsection{User Profiling}

User profiling is a core component of personalized services. 
Traditional machine learning approaches, such as regression or Bayesian models, treat each attribute (\textit{e.g.}, age, occupation) independently \cite{profile-age, profile-gender, profile-career}, limiting their ability to integrate heterogeneous behavioral signals and their reliance on labeled data.
With the rise of LLMs, recent work explores profile construction via prompting, fine-tuning \cite{profile-construct-update-benchmark,profile-youarewhatyoubought,profile-tuning-emb}, or synthetic data generation \cite{profile-construct-update-benchmark,profile-persona} to mitigate label scarcity. However, these methods often rely on LLM-generated labels without rigorous quality control and fail to capture the complexity of real-world user information. 
Beyond profile construction, several studies apply inferred profiles to downstream tasks such as recommendation or personalization. For instance, prior works leverage user embeddings or inferred demographic attributes to improve ranking models and alignment \cite{profile-youarewhatyoubought,profile-tuning-emb,profile-judge,profile-emb}. 
Nevertheless, these approaches typically evaluate only the downstream utility, rather than directly assessing the accuracy and robustness of the profiling process itself, restricting their broader applicability.
These limitations highlight the need for a more reliable framework that generates reliable profiles under label scarcity, enhances reasoning capability, and supports standardized evaluation with real-world user data.

\subsection{Confidence Utilization in LLMs} 

Confidence estimation is widely recognized as a key to improving the reliability of LLMs, as it enables models to quantify uncertainty. 
Early studies demonstrate that LLMs possess an inherent ability to estimate their own confidence \cite{conf-lm-knows, conf-survey}, and that prompting, sampling, or aggregation strategies can strengthen these signals \cite{conf-can-llm,ttrl}. 
To address the pervasive issue of overconfidence, subsequent work has proposed calibration strategies, including sampling-based calibration \cite{conf-tts-calibration}, structured prompting with distractors \cite{conf-mind-gap}, and likelihood-based alignment \cite{conf-towards-objective}. 
Recent research has further incorporated confidence into reasoning and reward design in Reinforcement Learning (RL). Examples include penalizing over- and under-confident predictions \cite{conf-reward-doubt}, maximizing distributional consistency \cite{conf-all-you-need}, and filtering low-quality reasoning trajectories based on confidence \cite{conf-deep-think}. 
However, these methods are primarily designed for mathematical and supervised reasoning tasks.
In contrast, \sysname~ targets the domain of user profiling, where acknowledging "unknown" is critical to improve factuality. We make full utilization of confidence across the entire pipeline, from data synthesis and distillation to reinforcement learning, achieving robust and label-free user profiling.

\section{ProfileBench}

To rigorously evaluate label-free user profiling, we construct ProfileBench, a novel benchmark derived from a large-scale industrial video platform.

\subsection{Data Collection and Anonymization}

ProfileBench is derived from a large-scale industrial video platform. Each user record integrates heterogeneous input signals from two sources:
\begin{itemize}
\item \textbf{Behavioral cues}, keyphrases summarizing the content of videos that a user has watched, searched, or submitted. These keyphrases are extracted from associated metadata and content analysis, forming a white-box textual representation of user interests. 
\item \textbf{Demographic cues}, publicly self-disclosed fields, such as age, gender, geographic region, and personal signature. These fields are often optional, incomplete, or noisy, making them unreliable as direct supervision signals.
\end{itemize}

To enable quantitative evaluation, we curate a human-annotated evaluation set of 1,000 users through standardized questionnaires. Participants were asked to verify their demographic information across all six dimensions. Inconsistent or implausible responses were manually filtered, resulting in a high-quality gold standard for benchmarking.

Only user-authorized data is included, with all information anonymized.
All user identifiers, nicknames, and URLs were removed during preprocessing to ensure privacy and compliance. 

\subsection{Multi-dimensional Tagging Taxonomy}

To provide a systematic framework for evaluation, we define a closed-set tagging taxonomy covering six fundamental dimensions:

\begin{itemize}
\item \textbf{Gender:} {Male, Female, Unknown}.
\item \textbf{Age:} {0–18, 18–23, 24–30, 31–40, 41–50, 50+, Unknown}
\item \textbf{Industry:} 14 categories (\textit{e.g.}, Manufacturing, Finance, Government, etc.), plus Unknown.
\item \textbf{Occupation:} 20 fine-grained categories (\textit{e.g.}, Students, Medical professionals, Retirees, Entrepreneurs, \textit{etc.}), plus Unknown. 
\item \textbf{Education Level:} {Junior high or below, Senior/vocational high, Bachelor/associate, Postgraduate or above, Unknown}.
\item \textbf{Life Stage:} 13 categories (\textit{e.g.}, Single, In a relationship, Parenting with different child age ranges), plus Unknown.
\end{itemize}

Each dimension can be formulated as a multi-class classification problem with an explicit Unknown (\texttt{NA}) option. 
This design acknowledges that user data can be sparse or ambiguous and avoids forcing the model to guess under high uncertainty. 
The inclusion of \texttt{NA} directly supports confidence-aware evaluation, enabling models to abstain when predictions are unreliable. 
The complete tagging taxonomy is detailed in Table \ref{tab:profile_system}. 

\subsection{Benchmark Characteristics and Challenges}

ProfileBench mirrors the complexities of the real world, while also presenting several unique challenges that make it a demanding benchmark for LLM-based profiling:
\begin{itemize}
    \item \textbf{Heterogeneous signals.} Each user record combines free-text behavioral keyphrases and noisy self-reported demographics, requiring models to integrate multi-sourced evidence.
    \item \textbf{Ambiguity and noise.} User behaviors may point to conflicting interests, and optional demographic fields are incomplete or outdated.
    \item \textbf{Multi-dimensional outputs.} The task requires jointly predicting six correlated attributes, rather than a single dimension, reflecting the real-world complexity.
    \item \textbf{Scarcity of ground-truths.} Except for the 1,000 annotated samples, the data lacks ground-truths, requiring profiling without explicit supervision.
\end{itemize}

Together, these factors ensure that ProfileBench is not only a practical dataset grounded in industry but also a challenging and realistic benchmark for advancing label-free user profiling.

\begin{figure*}[t]
  \centering
  \begin{subfigure}{.69\textwidth}
    \includegraphics[width=\linewidth]{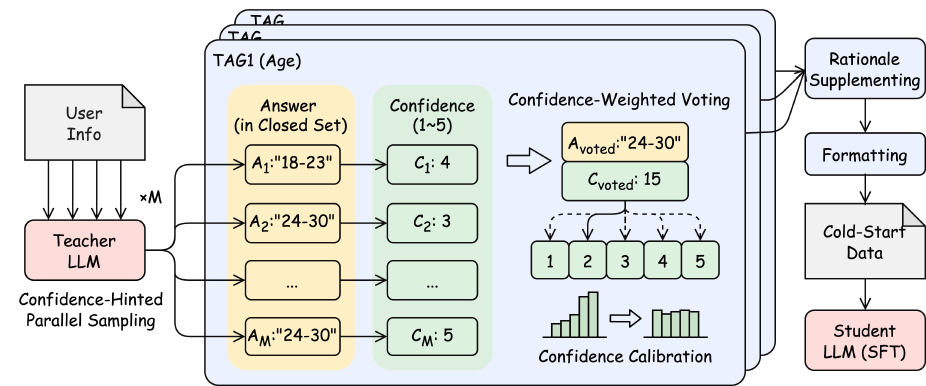}
    \caption{}
    \label{fig: data pipeline}
  \end{subfigure}
  \hfill 
  \begin{subfigure}{.28\textwidth}
    \includegraphics[width=\linewidth]{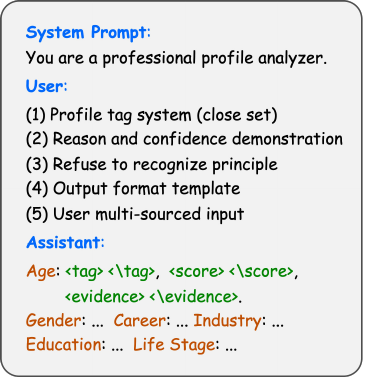}
    \caption{}
    \label{fig: prompt template}
  \end{subfigure}
  \caption{
    \textbf{The confidence-hinted cold-start data synthesis pipeline. }
    (a) Data synthesis. A teacher LLM is instructed to reason over multi-dimensional tags with confidence hints, followed by parallel sampling with confidence-weighted voting to improve the data quality, and confidence calibration to balance the confidence distribution. 
    (b) Prompt template. The constructed pseudo-labels, confidence scores, and evidence are formatted as SFT data, with each element enclosed within <box>...</box>.
    }
  \label{Fig2: data pipeline and template}
\end{figure*}

\section{Method}

\sysname~ features a two-stage unsupervised paradigm for profile reasoning: 1) cold-start data synthesis and distillation, and 2) reinforcement learning for further refinement, with explicit modeling and optimization of confidence throughout.

\subsection{Profiling Task Formulation}
\label{sec: Profiling Task Formulation}

Essentially, the user profiling task is defined as a multi-dimensional classification problem. 
Let $\mathcal{I}$ represent the collection of multi-source user information.
A set of attributes, $\mathcal{D} = \{D_1, \ldots, D_K\}$ are designed to portrait users. 
For attribute $D_k$, there is a predefined set of possible tags $\mathcal{T}_k = \{t_{k,1}, \ldots, t_{k,m}\} \cup \{\text{\texttt{NA}}\}$.
The objective is to learn a profiling model $f(\cdot)$ that maps the input information $\mathcal{I}$ to a specific tag $y_k \in \mathcal{T}_k$ for each attribute $D_k$:
\begin{equation}
    (y_1, y_2, \ldots, y_K) = f(\mathcal{I}).
\end{equation}

Notably, in cases where the input $\mathcal{I}$ offers insufficient evidence for a given attribute $D_k$, the model should learn to abstain from prediction by generating $y_k = \texttt{NA}$. 

\subsection{Confidence-Hinted Cold-Start Data Synthesis}
\label{sec: Data Synthesis}

Given the lack of ground-truths, we first harness a strong LLM to synthesize high-quality pseudo-labels, serving as the supervision for cold-start fine-tuning. The pipeline is shown in Figure \ref{Fig2: data pipeline and template}.

\textbf{Confidence-hinted Profiling.}
For a given user information $\mathcal{I}$, an advanced reasoning LLM (\textit{e.g.}, DeepSeek-R1), denoted as $f_{\text{teacher}}$, is prompted to generate a structured result for the $K$ attributes. 
To address the uncertainty stemming from sparse or ambiguous user information, we instruct the LLM to not only infer the profile tags but also provide the confidence scores and the supporting evidence for each tag. 
Each output thus forms a triplet consisting of the predicted tag $A_k$, a confidence score $C_k$, and the reasoning evidence $E_k$:
\begin{equation}
    \left\{ (A_k, C_k, E_k) \right\}_{k=1}^K = f_{\text{teacher}}(\mathcal{I}),
\end{equation}
where the confidence score $C_k$ is selected from a predefined 5-level verbal scale, ranging from low to high confidence.

Confidence constitutes a crucial component of the \sysname~ framework, underpinning threefold purposes.
1) It helps reduce hallucinations by compelling the model to assess the reliability of its own reasoning, thereby improving robustness.
2) It naturally reflects the inherent difficulty of certain attributes. High-confidence predictions are typically supported by explicit evidence, whereas low-confidence ones often correspond to speculative inferences. 
3) It enables a tunable balance between precision and recall. By setting a confidence threshold $\tau$, predictions with scores below it can be reverted to "\texttt{NA}", yielding higher precision at the cost of lower recall. 

\textbf{Parallel Sampling and Confidence-weighted Voting.} 
Single-round inference is vulnerable to sampling bias, which may lead to unreliable profile predictions. 
To mitigate this, for each user record $\mathcal{I}$, we perform $M$ times parallel inferences. 
The final profile is then determined via confidence-weighted voting, as illustrated in Figure \ref{fig: data pipeline}. 
For each attribute $D_k$, we aggregate the $M$ parallel predictions. 
The total confidence-weighted score for each candidate tag $t \in \mathcal{T}_k$ is calculated by summing up the numerical scores:

\begin{equation}
    S_{k,t} = \sum_{i=1}^{M} \mathbb{I}\!\left(A_{k,i} = t\right) \cdot C_{k,i},
\end{equation}

where $\mathbb{I}(\cdot)$ is an indicator function, 
$A_{k,i}$ denotes the predicted tag for attribute $D_k$ for the $i$-th sampling, and $C_{k,i}$ is the associated confidence score.

The voted result, $A_{k}^{\text{voted}}$, is the one with the highest aggregated score $S_k$:

\begin{equation}
    A_{k}^{\text{voted}} = \arg\max_{t \in \mathcal{T}_k} S_{k,t},
\end{equation}
\begin{equation}
    S_{k} = \max_{t \in \mathcal{T}_k} S_{k,t}.
\end{equation}

This strategy reduces sampling variance and ensures that the final decision is dominated by predictions with higher confidence, leading to more reliable and stable profiling results.

\begin{figure}[t]
  \centering
  \vspace{-0.5em}
  \begin{subfigure}{\columnwidth}
    \includegraphics[width=\linewidth]{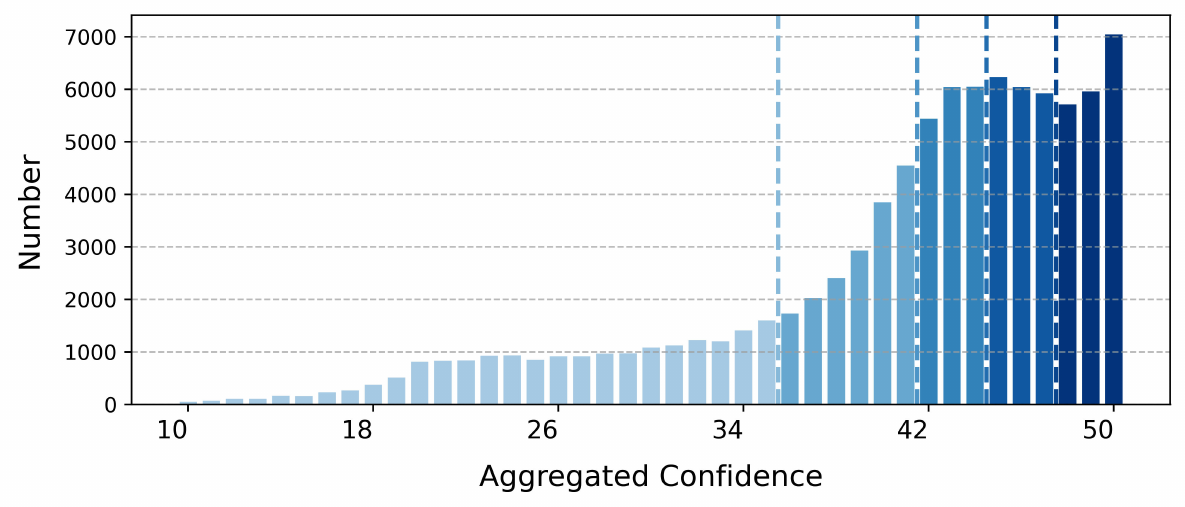}
    \caption{Gender}
    \label{Fig3a}
  \end{subfigure}
  \hfill
  \begin{subfigure}{\columnwidth}
    \includegraphics[width=\linewidth]{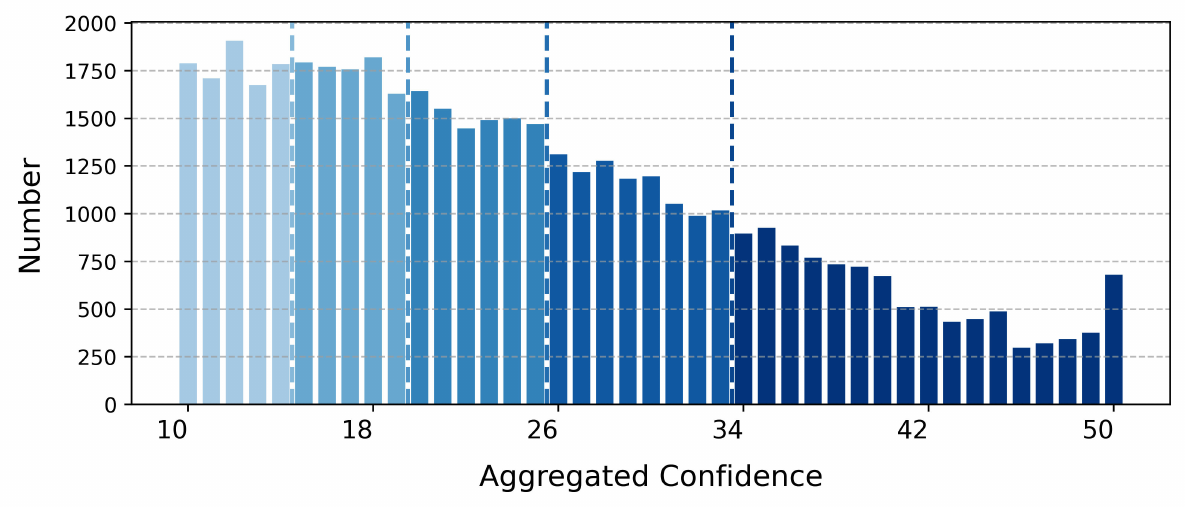}
    \caption{Life Stage}
    \label{Fig3b}
  \end{subfigure}
  \vspace{-1.6em}
  \caption{
    \textbf{The distribution and calibration of confidence. }
    The original aggregated confidence is unevenly distributed. 
    For instance, gender predictions tend to be highly confident, whereas life stage predictions are relatively under-confident. 
    The calibration step redistributes confidence values into five balanced intervals. 
    }
    \vspace{-0.1in}
  \label{fig: confidence calibration}
\end{figure}

\textbf{Confidence Calibration}. 
The raw confidence produced by the LLM is typically skewed and unevenly distributed, as illustrated in Figure~\ref{fig: confidence calibration}.
Such an imbalance occurs not only across different tags but also within the same tag, where confidence levels may be dispersed, leading to over-confident or under-confident predictions. 
To obtain a more balanced and informative signal for distillation, we calibrate the confidence to better align with the profiling task.
Specifically, the aggregated scores $S_k$ obtained from voting fall within a range of $[M, 5M]$. 
We partition this range into five intervals such that each interval contains approximately the same number of samples. 
This calibration process maps raw scores to a normalized confidence label $\widetilde{C}_k \in \{1, \dots, 5\}$, yielding a more stable and task-adaptive measure of prediction certainty.

\textbf{Synthetic Data Distillation.}
With the high-quality synthetic data generated, we proceed to distill the knowledge into a more lightweight student LLM, denoted as \( f_\theta \), via Supervised Fine-Tuning (SFT). 
To achieve this, we reformat the voted tags \( A_{k}^{\text{voted}} \), calibrated confidence scores \(\widetilde C_{k} \), and their corresponding rationales \( E_k \) into a structured format. 
As shown in Figure \ref{fig: prompt template}, each attribute's information is encapsulated within distinct XML-like tags: \texttt{<tag>}, \texttt{<score>}, and \texttt{<evidence>}.
This granular structure provides per-attribute rationales instead of a monolithic chain-of-thought. 
In this stage, the complex profiling capabilities of the teacher are transferred to the student, ensuring both adherence to structured output and consistency in reasoning. 
Moreover, the lightweight model enables better efficiency in both training and deployment.

\begin{figure*}[ht]
    \centering
    \includegraphics[width=\textwidth]{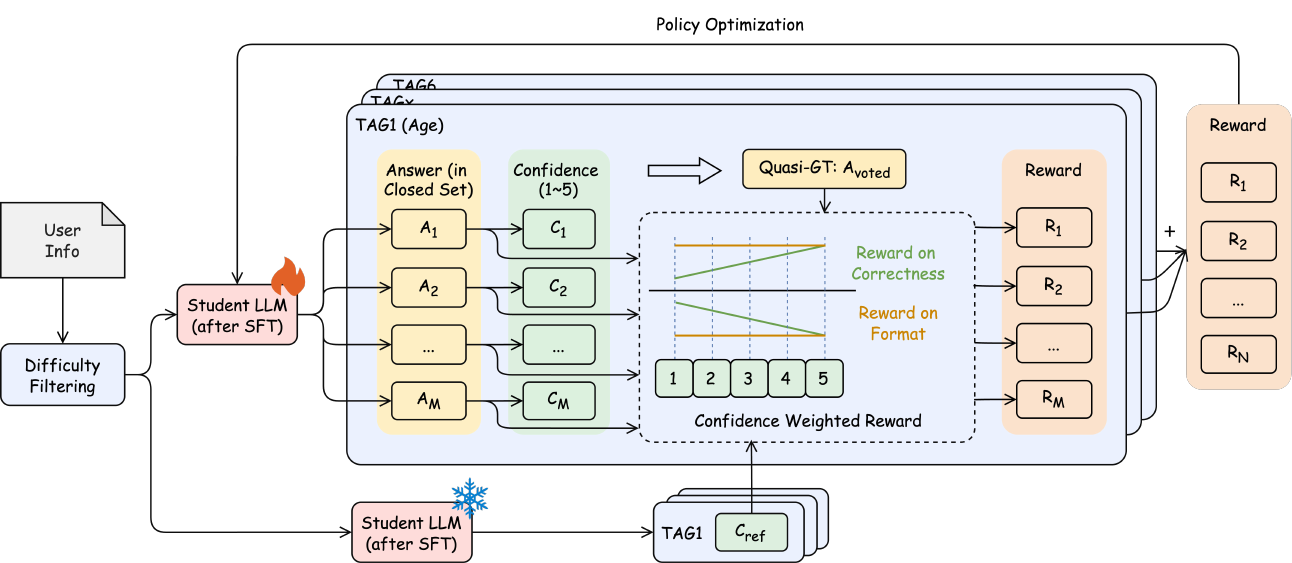}
    \caption{
    \textbf{The confidence-guided unsupervised reinforcement learning pipeline. }
    The pipeline is built on GRPO and incorporates confidence guidance. Given the label-free nature of the profiling task, the multiple rollouts are utilized to perform confidence-weighted voting, producing quasi-ground-truth labels. The reward function combines a format reward and a confidence-weighted reward, with frozen confidence values serving as weights. Rewards across multiple tags are summed to optimize the model.
    }
    \label{fig: rl}
\end{figure*}

\subsection{Confidence-Guided Unsupervised Reinforcement Learning}
\label{sec: Reinforcement Learning}

To further refine the reasoning capability of the distilled model, we employ online reinforcement learning based on GRPO \cite{shao2024deepseekmathpushinglimitsmathematical}, designing a confidence-guided reward to optimize without ground-truths.

\textbf{Confidence-based Difficulty Filtering.}
Difficulty filtering has proven to be an effective strategy for enhancing model capability. 
For the label-free profiling task, where it can be infeasible to rely on accuracy or human-labeled annotations for difficulty estimation, we adopt confidence as a natural proxy for sample difficulty. 
Concretely, we aggregate the confidence scores across multiple samplings of a sample and use it as the difficulty measurement. 
We then reshape the confidence distribution, such as filtering overly easy ones that provide little supervision or overly hard ones that may be hard to learn. 
This yields a more balanced set of samples that emphasizes informative, moderately difficult samples, ensuring richer learning signals and thereby improving the reasoning ability.

\textbf{Quasi-Groundtruth for Unsupervised RL.} 
A primary challenge in real-world profiling is the lack of ground-truth for reward calculation. 
To address this, we generate a high-quality optimization target, or "quasi-groundtruth", on the fly. 
For each input $\mathcal{I}$, we leverage the multiple rollouts and aggregate the resulting predictions using the same confidence-weighted voting mechanism described in Section \ref{sec: Data Synthesis}. 
For each attribute $D_k$, the quasi-groundtruth $A_{k, \text{quasi-GT}}$ is determined as follows:
\begin{equation}
    A_{k, \text{quasi-GT}} = \arg\max_{t \in \mathcal{T}_k} \sum_{m=1}^{M} \mathbb{I}(A_{k,m} = t) \cdot C_{k,m},
\end{equation}
where $(A_{k,m}, C_{k,m})$ is the prediction from the $m$-th rollout.
The voted tag, which represents the most consistent and confident inference of the current policy, serves as the optimization target for the correctness reward. 

\textbf{Confidence-Guided Reward.} 
To optimize for correctness, we introduce a reward that is further weighted by confidence. 
This mechanism is designed to make the model more sensitive to its own certainty, applying larger rewards or penalties to high-confidence predictions. 
The correctness reward is calculated by comparing the model's generated tag $A_k$ against the quasi-groundtruth $A_{k, \text{quasi-GT}}$:
\begin{equation}
    R_{k, \text{acc}} = w(C_{k, \text{ref}}) \cdot [2 \cdot \mathbb{I}(A_k = A_{k, \text{quasi-GT}}) - 1].
\end{equation}
where $w(\cdot)$ is a linear function of a frozen reference confidence $C_{k, \text{ref}}$, mapping confidence scores $\{1, ..., 5\}$ to weights $\{0.2, ..., 1.0\}$.
We use a frozen reference confidence rather than the online-generated confidence to avoid a phenomenon where the model might learn to generate artificially high confidence to maximize its potential reward. 
In practice, these reference scores are generated by a single offline inference pass of $f_{\theta}$ and remain fixed throughout RL training.

We also include a format reward to verify that the structural results can be successfully extracted and belong to the predefined closed sets. 
Let $O_k$ be the output for the the attribute $k$, and the format reward is defined as:
\begin{equation}
    R_{k} = 
    \begin{cases} 
    R_{k,acc} & \text{if } O_k \text{ is well-formed}, \\
    -1 & \text{otherwise}.
    \end{cases}
\end{equation}
The final reward for a rollout is the sum of all tags: $R = \sum_{k=1}^K R_{k}$.
This approach ensures that the model produces outputs that are both structurally valid and semantically accurate, while amplifying the predictions with higher confidence.

\section{Experiments}

We conduct extensive experiments to validate \sysname, comparing it with both SOTA LLMs and training-based baselines, and further analyzing the contribution of each component.

\subsection{Settings}

\textbf{Datasets}. 
We construct the training and evaluation datasets to reflect the unsupervised nature of real-world user profiling. 
For training, we collect 20,000 user samples without ground-truth labels. 
Among them, 10,000 samples are processed through the confidence-hinted data synthesis pipeline described in Section~\ref{sec: Data Synthesis}, providing high-quality supervision for SFT. 
The remaining 10,000 samples are used in the reinforcement learning stage described in Section~\ref{sec: Reinforcement Learning}.
For evaluation, we adopt ProfileBench, which provides heterogeneous user cues and a structured profiling system. 

\textbf{Model Configuration}. 
We employ DeepSeek-R1 \cite{deepseekai2025deepseekr1incentivizingreasoningcapability} via one-shot prompting to synthesize high-quality supervision data and set the parallel sampling number to 10.
We use Qwen3-8B \cite{yang2025qwen3technicalreport} as the base model, undergoing a two-stage training.
For the SFT stage, we use 10,000 synthesized samples, with full-parameter tuning. 
The learning rate, batch size, and training epochs are set to 1e-5, 32, and 2, respectively.
For the RL stage, we employ GRPO \cite{shao2024deepseekmathpushinglimitsmathematical} based on the VeRL framework \cite{sheng2024hybridflow}. 
We set batch size to 128, KL coefficient to 0.001, rollout number to 32, lr to 1e-6, sampling temperature to 1, top-p to 0.95, and employ 45 training steps.

\textbf{Baselines}. 
We compare \sysname~ against three categories of baselines. 
The first category involves prompting of SOTA reasoning LLMs (\textit{e.g.} Kimi-K2 \cite{kimiteam2025kimik2openagentic}, Seed-1.6-thinking \cite{Seed1.6}, and DeepSeek-R1 \cite{deepseekai2025deepseekr1incentivizingreasoningcapability}) to infer over single attribute or multiple attributes, serving as a measure of raw inference capability. 
The second is traditional machine learning models (\textit{e.g.} XGBoost \cite{Chen_2016}, LightGBM \cite{NIPS2017_6449f44a}) trained separately for each attribute, serving as strong legacy baselines.
The third is fine-tuned LLMs, represented by Profile-LLM, which is trained on the same unsupervised corpus as our method but without the confidence-hinted data synthesis or confidence-guided reinforcement learning, providing a vanilla fine-tuning baseline.

\textbf{Evaluation Metrics}. Given the unsupervised nature of the dataset, we evaluate model performance on the annotated evaluation set using recall, precision, and F1 score. 
To mitigate potential fluctuations, all reported results derived from the LLM are presented as the average of ten runs.

\begin{table*}[ht]
\centering
\setlength{\tabcolsep}{6pt}
\resizebox{\textwidth}{!}{
\begin{tabular}{cc|cccccc|c}
\toprule
\multicolumn{2}{c|}{\textbf{Models}} & \textbf{Age} & \textbf{Gender} & \textbf{Industry} & \textbf{Occupation} & \textbf{Education Level} & \textbf{Life Stage} & \textbf{Average} \\
\midrule
\multirow{6}{*}{SOTA LLMs}  
&     \cellcolor{brown!10}{Kimi-K2$^\text{\dag}$} & \cellcolor{brown!10}{78.00} & \cellcolor{brown!10}{94.01} & \cellcolor{brown!10}{63.36} & \cellcolor{brown!10}{62.74} & \cellcolor{brown!10}{52.27} & \cellcolor{brown!10}{57.25} & \cellcolor{brown!10}{67.94} \\
&     \cellcolor{brown!10}{Seed1.6-thinking$^\text{\dag}$} & \cellcolor{brown!10}{84.29} & \cellcolor{brown!10}{91.79} & \cellcolor{brown!10}{63.87} & \cellcolor{brown!10}{58.43} & \cellcolor{brown!10}{51.18} & \cellcolor{brown!10}{55.08} & \cellcolor{brown!10}{67.44} \\
&     \cellcolor{brown!10}{DeepSeek-R1$^\text{\dag}$} & \cellcolor{brown!10}{83.66} & \cellcolor{brown!10}{92.60} & \cellcolor{brown!10}{62.32} & \cellcolor{brown!10}{33.76} & \cellcolor{brown!10}{52.40} & \cellcolor{brown!10}{52.08} & \cellcolor{brown!10}{62.80} \\
&     \cellcolor{yellow!10}{Kimi-K2} & \cellcolor{yellow!10}{82.05} & \cellcolor{yellow!10}{91.43} & \cellcolor{yellow!10}{63.83} & \cellcolor{yellow!10}{63.77} & \cellcolor{yellow!10}{57.48} & \cellcolor{yellow!10}{45.42} & \cellcolor{yellow!10}{67.33} \\
&     \cellcolor{yellow!10}{Seed1.6-thinking} & \cellcolor{yellow!10}{83.98} & \cellcolor{yellow!10}{95.86} & \cellcolor{yellow!10}{67.37} & \cellcolor{yellow!10}{65.21} & \cellcolor{yellow!10}{55.50} & \cellcolor{yellow!10}{35.84} & \cellcolor{yellow!10}{67.29} \\
&     \cellcolor{yellow!10}{DeepSeek-R1} & \cellcolor{yellow!10}{83.76} & \cellcolor{yellow!10}{96.25} & \cellcolor{yellow!10}{67.89} & \cellcolor{yellow!10}{63.24} & \cellcolor{yellow!10}{51.29} & \cellcolor{yellow!10}{46.03} & \cellcolor{yellow!10}{68.08} \\
\midrule
\multirow{7}{*}{\shortstack{Training-based}}  
&     \cellcolor{brown!13}{ML-based$^\text{\dag}$} & \cellcolor{brown!13}{78.46} & \cellcolor{brown!13}{94.41} & \cellcolor{brown!13}{60.11} & \cellcolor{brown!13}{52.63} & \cellcolor{brown!13}{51.74} & \cellcolor{brown!13}{39.74} & \cellcolor{brown!13}{62.85} \\
&     \cellcolor{orange!10}{Qwen3-8B} (base model) & \cellcolor{orange!10}{84.40} & \cellcolor{orange!10}{95.64} & \cellcolor{orange!10}{51.10} & \cellcolor{orange!10}{51.31} & \cellcolor{orange!10}{40.9} & \cellcolor{orange!10}{31.03} & \cellcolor{orange!10}{59.06} \\
&     \cellcolor{orange!10}{Profile-LLM (\textit{w.} SFT)} & \cellcolor{orange!10}{83.89} & \cellcolor{orange!10}{95.72} & \cellcolor{orange!10}{65.33} & \cellcolor{orange!10}{61.14} & \cellcolor{orange!10}{52.08} & \cellcolor{orange!10}{44.11} & \cellcolor{orange!10}{67.05} \\
&     \cellcolor{orange!10}{Profile-LLM (\textit{w.} SFT+RL)} & \cellcolor{orange!10}{83.90} & \cellcolor{orange!10}{95.73} & \cellcolor{orange!10}{65.15} & \cellcolor{orange!10}{60.79} & \cellcolor{orange!10}{50.86} & \cellcolor{orange!10}{42.62} & \cellcolor{orange!10}{66.51} \\
&     \cellcolor{red!10}{\sysname~(\textit{w.} SFT)} & \cellcolor{red!10}{83.79} & \cellcolor{red!10}{96.26} & \cellcolor{red!10}{68.02} & \cellcolor{red!10}{63.03} & \cellcolor{red!10}{55.54} & \cellcolor{red!10}{51.35} & \cellcolor{red!10}{69.67} \\
&     \cellcolor{red!10}{\sysname~(\textit{w.} RL)} & \cellcolor{red!10}{84.41} & \cellcolor{red!10}{95.77} & \cellcolor{red!10}{51.04} & \cellcolor{red!10}{56.10} & \cellcolor{red!10}{23.37} & \cellcolor{red!10}{28.81} & \cellcolor{red!10}{56.58} \\
&     \cellcolor{red!10}{\sysname~(\textit{w.} SFT+RL)} & \cellcolor{red!10}{\textbf{84.54}} & \cellcolor{red!10}{\textbf{96.29}} & \cellcolor{red!10}{\textbf{71.63}} & \cellcolor{red!10}{\textbf{69.68}} & \cellcolor{red!10}{\textbf{58.55}} & \cellcolor{red!10}{\textbf{57.52}} & \cellcolor{red!10}{\textbf{73.03}} \\
\bottomrule
\end{tabular}
}
\caption{
\textbf{The performance comparisons of different methods.}  
$^\dag$ indicates single-dimensional inference. 
We report the F1 score and the best results are highlighted with \textbf{bold}.
\sysname~outperforms the raw capabilities of SOTA LLMs, as well as both traditional ML-based models and directly fine-tuned LLMs (Profile-LLM).
}
\label{tab:model_performance}
\end{table*}

\begin{figure}[t]
    \centering
    \subfloat[Precision across confidence thresholds.] {\includegraphics[width=1.0\columnwidth]{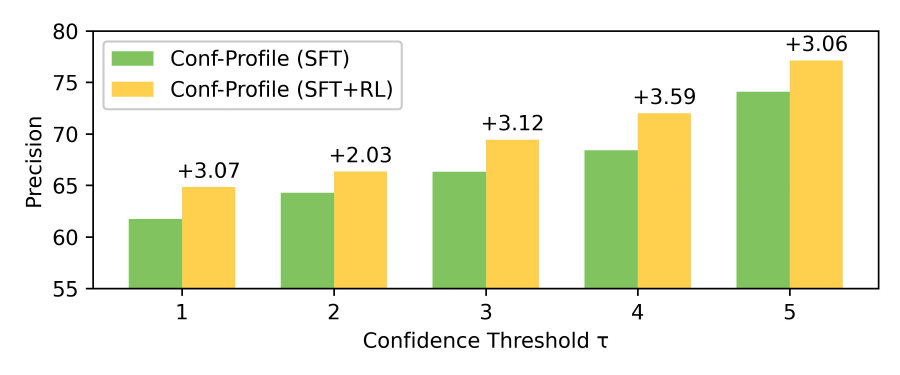}\vspace{-0.5em}\label{fig_attribution_1}}
    \hfill
    \subfloat[Recall across confidence thresholds.]{\includegraphics[width=1.0\columnwidth]{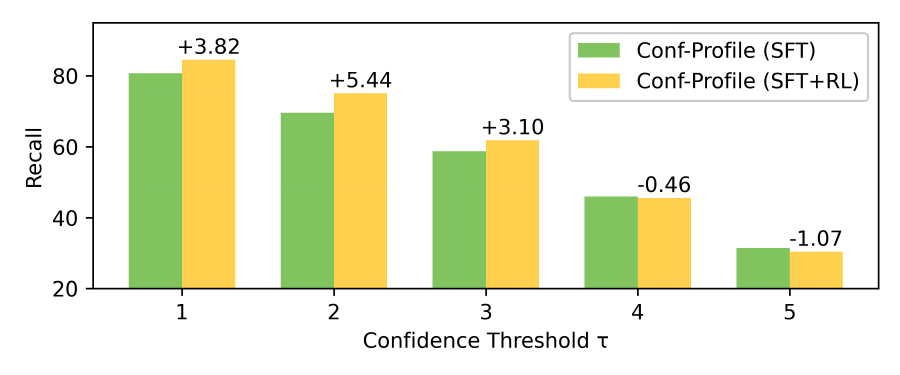}\vspace{-0.5em}\label{fig_attribution_3}}
    \caption{
    \textbf{The effects of confidence.}
    By setting a confidence threshold $\tau$, where predictions below it are set as '\texttt{NA}', we obtain multiple precision-recall trade-offs. 
    A higher $\tau$ corresponds to higher confidence, resulting in higher precision and lower recall. 
    }
    \label{fig: tradeoff}
\end{figure}

\subsection{Main Results}

Table~\ref{tab:model_performance} summarizes the F1 scores of prompt-based baselines, training-based baselines, and our proposed \sysname. 
We highlight three key observations.
1) SOTA LLMs achieve decent performance and outperform traditional ML-based models.
Among them, DeepSeek-R1 with multi-dimensional joint inference attains the highest F1 score, motivating its use as the teacher model. 
2) Fine-tuning the base model without confidence offers limited gains: SFT improves moderately but remains 2.52 points below \sysname, and SFT+RL without confidence can even degrade performance, underscoring the importance of confidence guidance.
3) \sysname~ delivers substantial improvements through two-stage training: SFT raises the average F1 by 10.61 points, and subsequent RL boosts 3.36 points. Direct RL on the base model fails to improve results, highlighting the need for cold-start fine-tuning.
Overall, \sysname~ achieves the best performance across all six evaluation dimensions, with an average F1 of 73.03, demonstrating its superiority and robustness for multi-dimensional user profiling.

We further validate the role of confidence by analyzing the impact of a rejection threshold $\tau$.
By treating predictions with confidence $C < \tau$ as \texttt{NA}, we obtain hierarchical precision–recall trade-offs. 
As illustrated in Figure \ref{fig: tradeoff}, high thresholds correspond to high precision, confirming that confidence serves as an effective indicator of prediction difficulty. 
Notably, after RL, the performance curve improves consistently across all thresholds, underscoring the value of confidence-guided optimization.

\subsection{Ablation Study}

We ablate the effects of confidence utilization in different stages, including aggregation, calibration, filtering, and reward guidance.

\textbf{Aggregation for Parallel Sampling.} 
We utilize various aggregation strategies during parallel sampling to produce quasi-labels. 
Table \ref{tab:voting_performance} reports the resulting SFT performance on ProfileBench.
Majority voting slightly improves precision but reduces recall due to equal voting weights on \texttt{NA} predictions, which correspond to low confidence. 
Confidence-weighted voting down-weights low-confidence outputs, improving both precision and recall, and achieves an F1 gain of 2.87.

\textbf{Confidence Calibration.} 
On the one hand, Figure \ref{fig: confidence calibration} clearly demonstrates that confidence calibration produces a more uniform data distribution.
On the other hand, we conduct experiments on SFT and RL without calibration. 
As shown in Table \ref{tab: calibration}, training with calibrated data improves the F1 score by 0.23 after SFT and by 0.26 after SFT+RL, respectively. 
This demonstrates that calibration not only yields a more balanced distribution but also helps improve the final performance.

\textbf{Difficulty Filtering for RL.} 
We investigate the impact of adjusting training data distribution by filtering samples according to their confidence. 
As shown in Table~\ref{tab:data_filtering}, reshaping the distribution significantly influences performance and enables a controllable precision–recall adjustment. 
The $\vee$-shape distribution achieves the highest recall, while the M-shape distribution yields the best precision. 
This is intuitive: filtering more low-confidence samples reduces \texttt{NA} outputs, whereas filtering medium-difficulty samples enhances the reliability of aggregation. 
Since precision is generally harder to enhance, we report the M-shape as our default configuration.

\textbf{Confidence Guiding in RL.} 
We further compare the source of confidence used for reward signals, as shown in Table~\ref{tab: frozen confidence}. 
With self-generated confidence, the model quickly learns to inflate its scores toward high values, leading to minimal variation across thresholds. 
This results in higher recall but undermines the calibration role of confidence. 
In contrast, using the frozen confidence preserves the meaningful calibration, maintaining a balanced precision–recall curve across thresholds and thus yielding more reliable supervision.

\begin{table}[t]
\centering
\resizebox{\columnwidth}{!}{
\begin{tabular}{c|ccc}
\toprule
\textbf{Metrics} & \hspace{8pt}\textbf{Mean}\hspace{8pt}  & \textbf{Majority-Voted} & \textbf{Confidence-Voted} \\
\midrule
Precision & 61.62 & \textbf{62.00}   & 61.76     \\
Recall    & 74.41 & 73.33   & \textbf{80.69}     \\
\cellcolor{gray!10}{F1 Score}  & \cellcolor{gray!10}{67.10} & \cellcolor{gray!10}{67.17}   & \cellcolor{gray!10}{\textbf{69.97}}     \\
\bottomrule
\end{tabular}
}
\caption{
\textbf{Comparisons of voting strategies for parallel sampling.}
Confidence-voted aggregation achieves the highest F1 score.
}
\label{tab:voting_performance}
\end{table}

\begin{table}[t]
\centering
\resizebox{\columnwidth}{!}{
    \begin{tabular}{c|ccccc}
    \toprule
    \textbf{Metrics} & 
    \makecell{\textbf{Original} \\ \includegraphics[width=1cm]{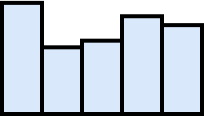}} & 
    \makecell{\textbf{Uniform} \\ \includegraphics[width=1cm]{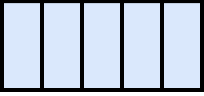}} & 
    \makecell{\textbf{\(\vee\)-Shape} \\ \includegraphics[width=1cm]{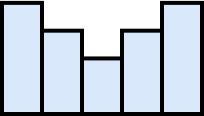}} & 
    \makecell{\textbf{\(\wedge\)-Shape} \\ \includegraphics[width=1cm]{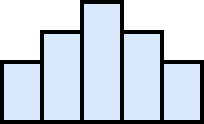}} & 
    \makecell{\textbf{M-Shape} \\ \includegraphics[width=1cm]{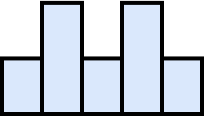}} \\
    \midrule
    Precision & 63.30 & 63.14 & 63.84 & 62.25 & \textbf{64.83} \\
    Recall    & 80.97 & 87.49 & 85.95 & \textbf{90.19} & 84.51 \\
    \cellcolor{gray!10}{F1 Score}  & \cellcolor{gray!10}{70.68} & \cellcolor{gray!10}{72.89} & \cellcolor{gray!10}{72.94} & \cellcolor{gray!10}{73.02} & \cellcolor{gray!10}{\textbf{73.03}} \\
    \bottomrule
    \end{tabular}
}
\caption{
\textbf{Comparisons of data filtering shape.}
\(\wedge\)-shape yields highest recall, while M-shape yields highest precision.
}
\label{tab:data_filtering}
\end{table}

\begin{table}[t]
\centering
\resizebox{\columnwidth}{!}{
\begin{tabular}{c|cc|cc}
\toprule
\multirow{2}{*}{\textbf{Metrics}} & \multicolumn{2}{c}{\textbf{\makecell{\sysname~ (\textit{w.o.} calibration)}}} & \multicolumn{2}{|c}{\textbf{\sysname}} \\
\cmidrule(lr){2-3} \cmidrule(lr){4-5}
                         & \textbf{SFT}       & \textbf{SFT+RL}       & \textbf{SFT}       & \textbf{SFT+RL}       \\
\midrule
Precision                & 61.51     & 64.42        & 61.76     & 64.83        \\
Recall                   & 80.52     & 84.52        & 80.69     & 84.51        \\
\cellcolor{gray!10}{F1 Score}                 & \cellcolor{gray!10}{69.74}     & \cellcolor{gray!10}{73.11}        & \cellcolor{gray!10}{\textbf{69.97}}     & \cellcolor{gray!10}{\textbf{73.37}}        \\
\bottomrule
\end{tabular}
}
\caption{
\textbf{Impact of confidence calibration. }
By incorporating confidence calibration, the F1 score increases by 0.23 and 0.26 after SFT and SFT+RL, respectively.
}
\label{tab: calibration}
\end{table}

\begin{table}[t]
\centering
\setlength{\tabcolsep}{13pt}
\resizebox{\columnwidth}{!}{
\begin{tabular}{c|cc|cc}
\toprule
\multirow{2}{*}{\makecell{\textbf{$\tau$}}} & \multicolumn{2}{c|}{\textbf{Self-Confidence}} & \multicolumn{2}{c}{\textbf{Frozen Confidence}} \\
& \textbf{Precision} & \textbf{Recall} & \textbf{Precision} & \textbf{Recall} \\
\midrule
1 & 61.38 & 95.05 & 64.83 & 84.51 \\
2 & 61.48 & 94.83 & 66.34 & 75.06 \\
3 & 61.57 & 94.54 & 69.44 & 61.83 \\
4 & 61.58 & 94.32 & 72.01 & 45.53 \\
5 & 61.94 & 91.90 & 77.14 & 30.35 \\
\bottomrule
\end{tabular}
}
\caption{\textbf{Comparisons of confidence guidance.}
Self-confidence voting leads to minimal variation under different confidence, while frozen confidence can maintain the hierarchy confidence distribution.
}
\label{tab: frozen confidence}
\end{table}

\section{Conclusion}

In this paper, we present \sysname, a comprehensive framework that advances label-free user profiling by leveraging confidence as both supervision and optimization signals. 
Our contributions include ProfileBench, the first industry-level user profiling benchmark grounded on real-world scenarios, a confidence-hinted data synthesis pipeline for generating high-quality cold-start supervision, and a confidence-guided unsupervised reinforcement learning scheme to strengthen reasoning ability. 
Together, these components deliver significant gains on profiling performance.
We believe this work not only establishes a strong foundation for user profiling but also opens promising directions for applying LLMs to broader user modeling tasks.

\section{Limitations}

While \sysname~ substantially advances confidence-driven user profiling, several aspects remain open for further exploration. 
Currently, ProfileBench focuses on several close-set dimensions, which provide a practical starting point but fall short of capturing the fine-grained user attributes. 
Extending the benchmark toward more granular, open-set profiling with dedicated evaluation protocols would better reflect real-world demands.
Moreover, user behavior histories might be extremely long and noisy in practice, posing challenges for reasoning over ultra-long, heterogeneous contexts. 
Finally, although our confidence-guided reinforcement learning demonstrates strong potential in fully unsupervised settings, incorporating hybrid supervision, where limited human annotations complement unsupervised optimization could further enhance robustness and interpretability.

\section{Ethics}

Given potential fairness concerns in algorithmic inference, all data used in this work were collected from users who explicitly consented to their information being used for research purposes. 
Personally identifiable information was removed or anonymized prior to processing.
Sensitive data, such as consumption behavior, is handled carefully to guarantee that no raw personal data are utilized. 
Our framework emphasizes uncertainty modeling and confidence-aware reasoning to mitigate the risks of incorrect profiling. 
All predictions are accompanied by confidence scores, with low-confidence outputs flagged for potential human review. 
The study was conducted following ethical research guidelines, and all procedures were internally reviewed to ensure compliance with privacy standards. 
The proposed methods are intended for research and responsible application in settings where user consent is guaranteed, and they provide only probabilistic predictions at the population level.

\bibliography{custom}

\appendix
\clearpage
\begin{table*}[htbp]
  \centering
  \setlength{\tabcolsep}{6pt}
  \begin{tabular*}{\linewidth}{@{\extracolsep{\fill}} p{1.3cm} p{0.7cm} p{11cm}@{}}
    \toprule
    \textbf{Dimension} & \textbf{Count} & \textbf{Category} \\
    \midrule
    Gender & 3 & "Female", "Male", "Unknown(NA)" \\
    \addlinespace
    Age & 7 & "0-18", "18–23", "24–30", "31–40", "41–50", "50+", "Unknown(NA)" \\
    \addlinespace
    Industry & 14 & "Agriculture and Fishery", "Manufacturing", "Real Estate and Construction", "Commerce and Retail", "Transport and Logistics", "High-Tech", "Services", "Finance", "Education and Training", "Healthcare", "Media, Culture, Sports and Entertainment", "Government and Public Institutions", "Not Employed", "Unknown(NA)" \\
    \addlinespace
    Occupation & 20 & "Software", "Clerical Staff", "Education and Trainer", "Beauty and Hairdressing", "Skilled Workers", "Government and Public Sector", "Transportation and Logistics", "Hospitality and Entertainment", "Media and Culture", "Independent Media", "Healthcare", "Agriculture and Fishery", "Finance and Insurance", "Self-Employed", "Domestic and Security", "Student", "High-Tech Hardware", "Retiree", "Homemaker", "Unknown(NA)" \\
    \addlinespace
    Life Stage & 13 & "Single", "In Relationship", "Pre-Marital", "Married, No Children", "Pre-Pregnancy and Pregnancy", "Parenting (Child 0–2)", "Parenting (Child 3–5)", "Parenting (Child 6–11)", "Parenting (Child 12–14)", "Parenting (Child 15–17)", "Parenting (Adult Child)",  "Parenting (Child Age Unknown)", "Unknown(NA)" \\
    \addlinespace
    Education Level & 5 & "Junior High or Below", "Senior/Vocational High", "Bachelor/Associate", "Postgraduate or Above", "Unknown(NA)" \\
    \bottomrule
  \end{tabular*}
  \caption{Detailed profiling tagging taxonomy.} 
  \label{tab:profile_system} 
\end{table*}

\begin{figure*}[ht]
  \centering
  \begin{subfigure}{.49\textwidth}
    \includegraphics[width=\linewidth]{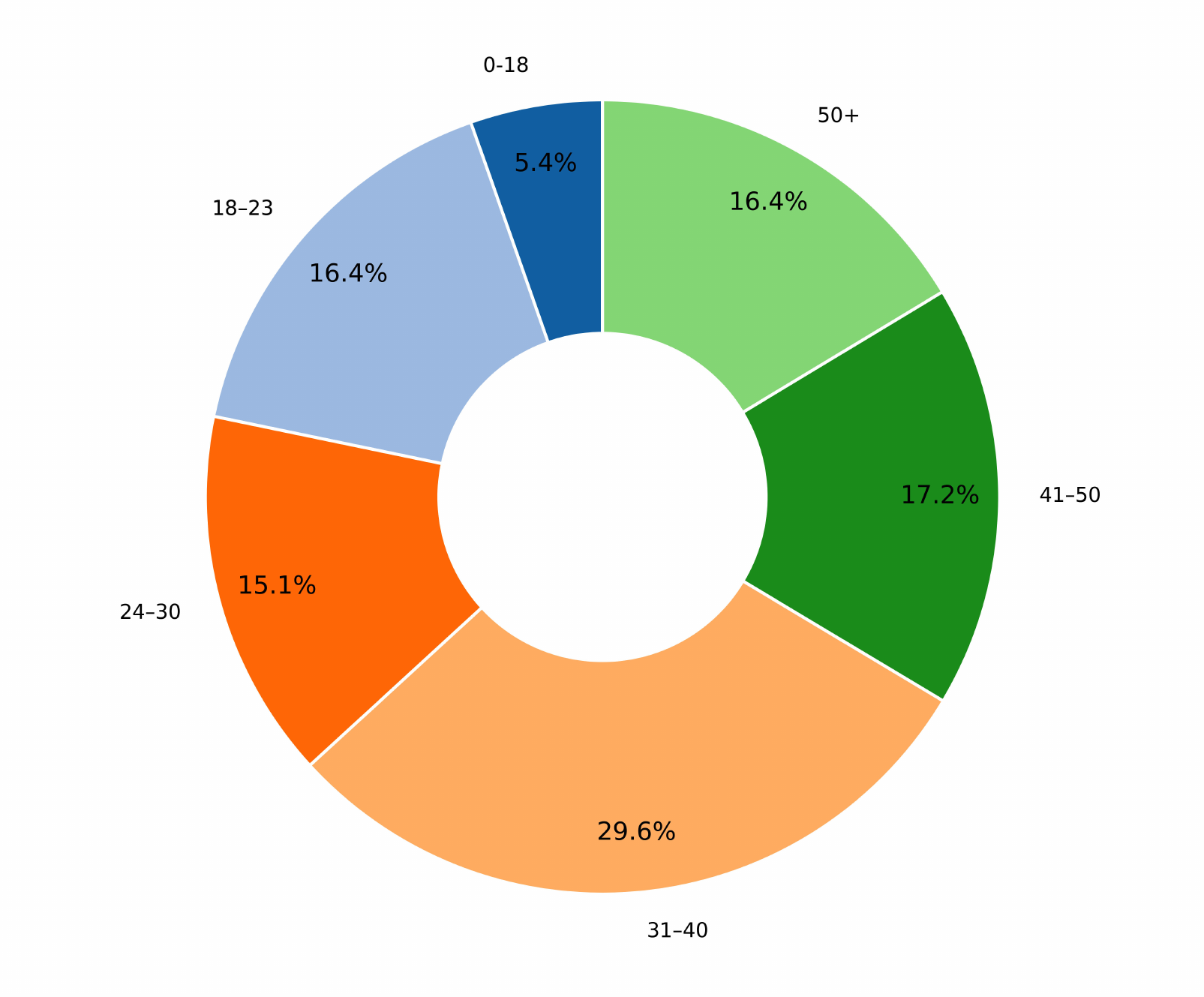}
    \caption{Distribution of Age}
    \label{fig: dis_age}
  \end{subfigure}
  \hfill 
  \begin{subfigure}{.49\textwidth}
    \includegraphics[width=\linewidth]{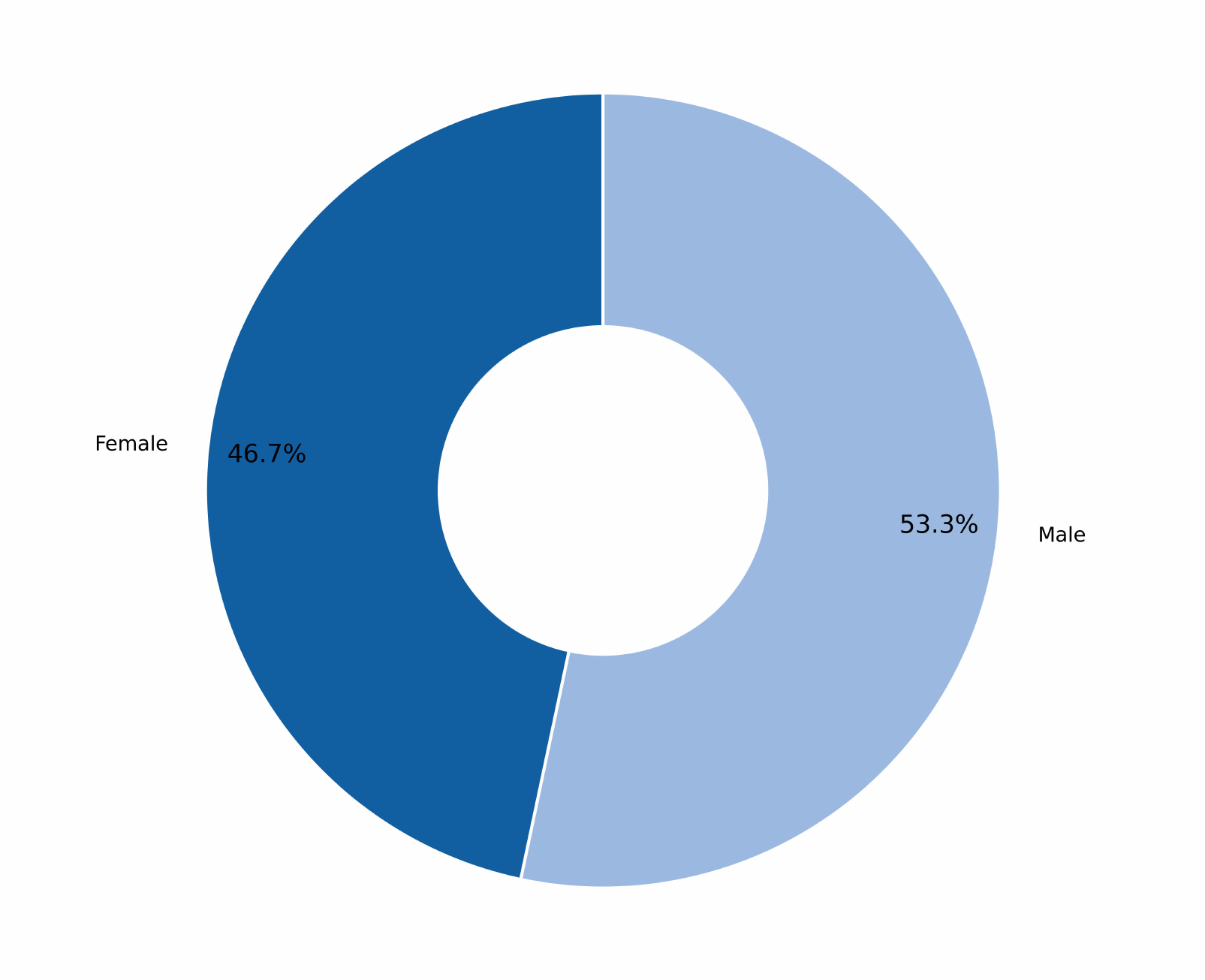}
    \caption{Distribution of Gender}
    \label{fig: dis_gender}
  \end{subfigure}
  \hfill 
  \begin{subfigure}{.49\textwidth}
    \includegraphics[width=\linewidth]{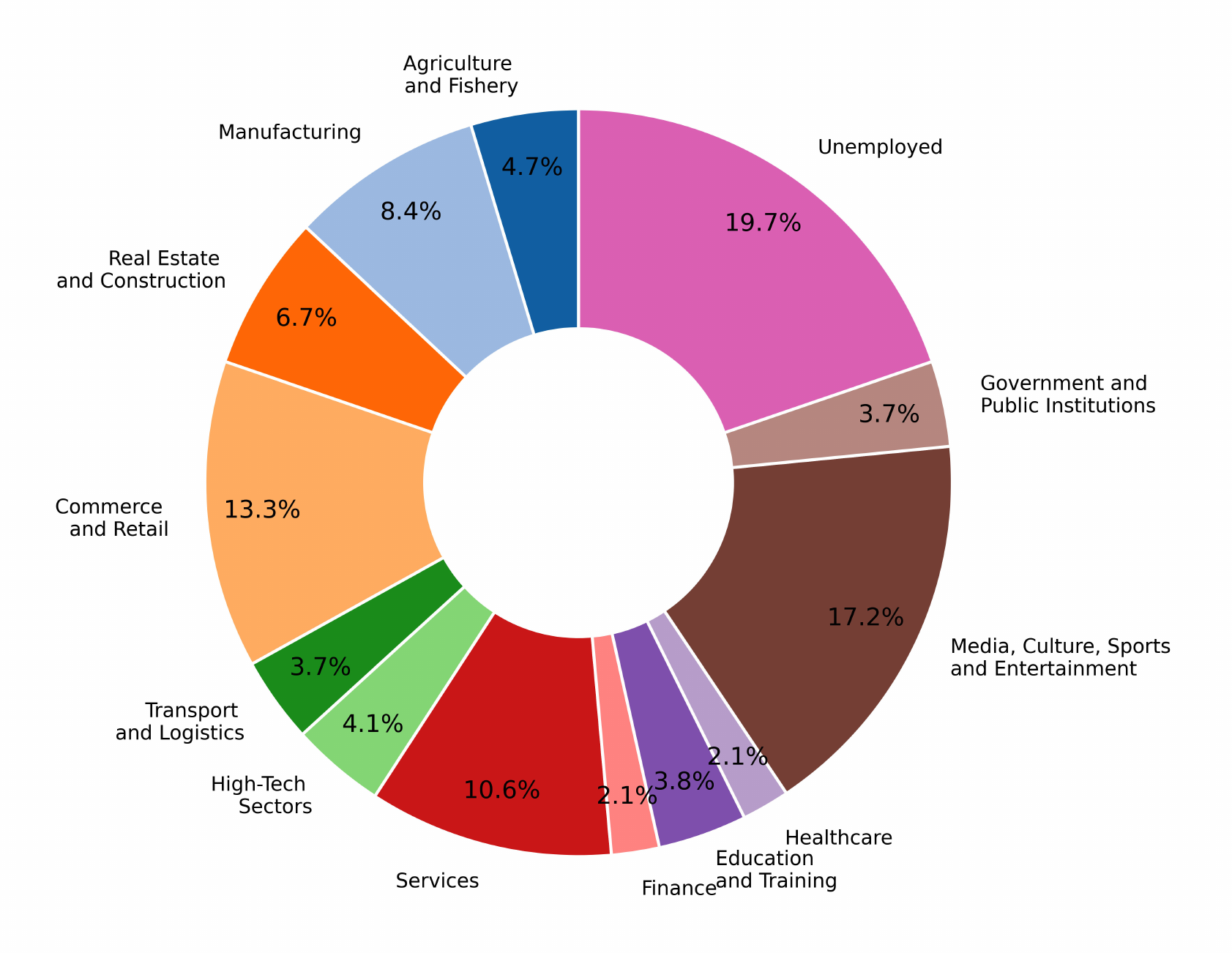}
    \caption{Distribution of Industry}
    \label{fig: dis_industry}
  \end{subfigure}
  \hfill 
  \begin{subfigure}{.49\textwidth}
    \includegraphics[width=\linewidth]{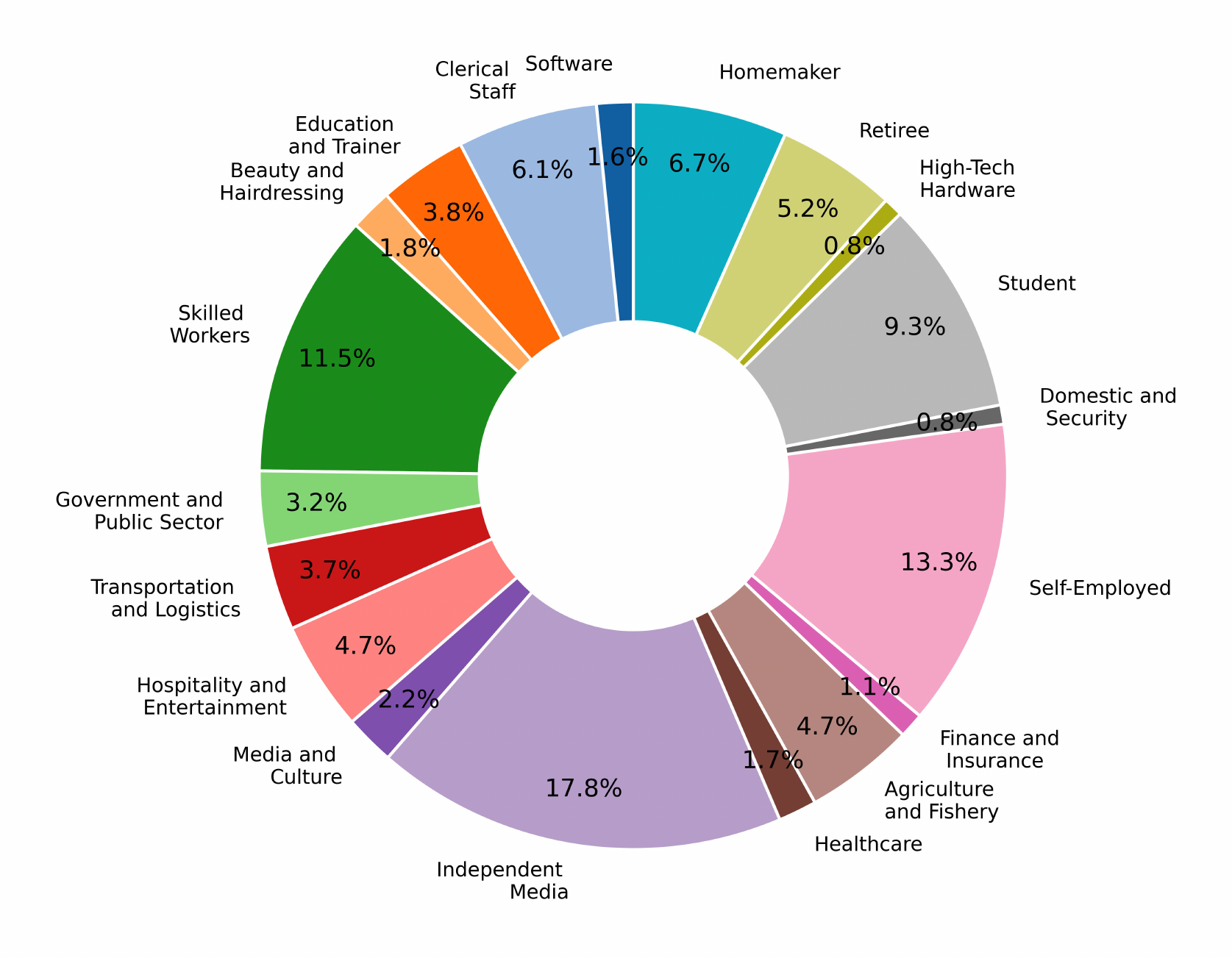}
    \caption{Distribution of Occupation}
    \label{fig: dis_occupation}
  \end{subfigure}
  \hfill 
  \begin{subfigure}{.49\textwidth}
    \includegraphics[width=\linewidth]{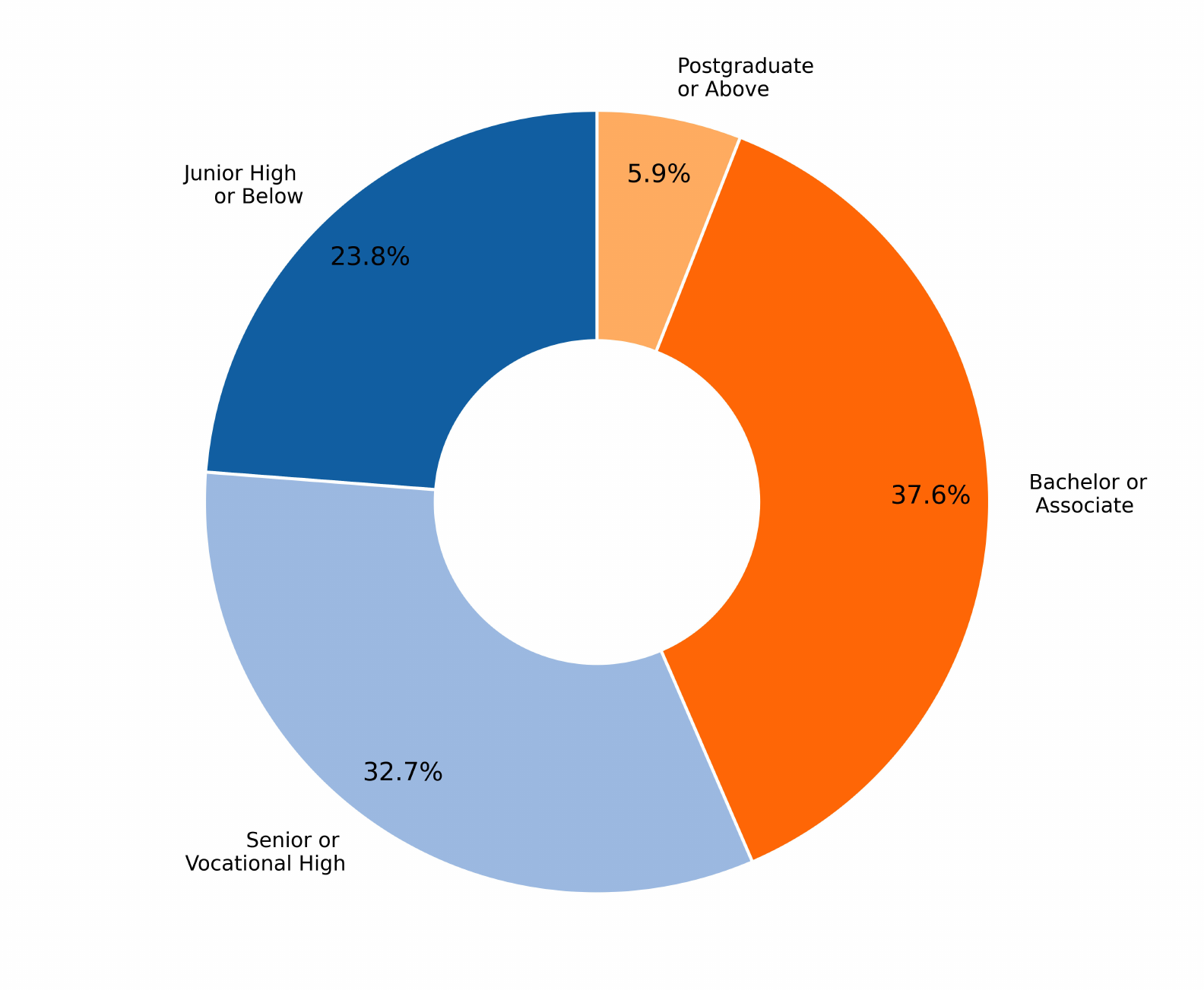}
    \caption{Distribution of Education}
    \label{fig: dis_education}
  \end{subfigure}
  \hfill 
  \begin{subfigure}{.49\textwidth}
    \includegraphics[width=\linewidth]{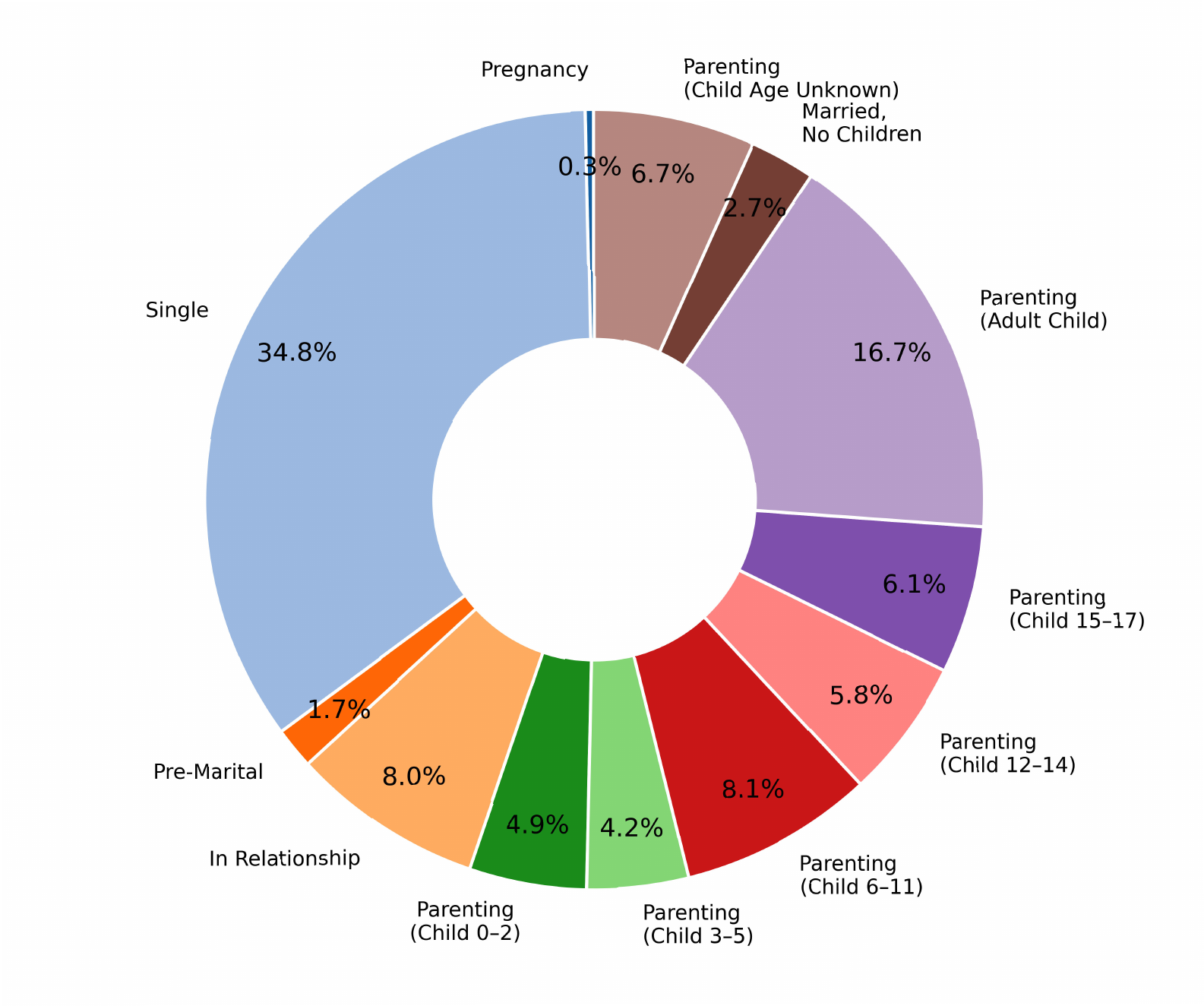}
    \caption{Distribution of Life Stage}
    \label{fig: dis_life}
  \end{subfigure}
  \caption{
    The distribution of ProfileBench across different attribute dimensions.
    }
  \label{fig: distribution ring}
\end{figure*}

\section{Appendix}

\subsection{Profile Tagging Taxonomy}
We present the complete profiling system in Table~\ref{tab:profile_system}, which spans six core dimensions, including gender, age, industry, occupation, education level, and life stage. 
The profiling system adopts a closed-set label taxonomy for each dimension, which facilitates consistent evaluation by ensuring that model predictions can be directly compared against standardized ground truth labels. 
Together, these attributes provide a comprehensive representation of user characteristics and serve as the foundation for user understanding. 

\subsection{ProfileBench Distribution}

We provide a high-quality benchmark, ProfileBench, consisting of 1k carefully curated samples derived from questionnaire-based data with human verification. 
Figure~\ref{fig: distribution ring} illustrates the distribution of data across different attribute dimensions in ProfileBench with ring charts. 
The samples are drawn from a large-scale video platform, ensuring a representative coverage of real-world distributions. 
In particular, for attributes with larger category spaces, such as occupation and industry, the test set maintains broad coverage to guarantee fairness and robustness in evaluation. 
Overall, ProfileBench offers a reliable, representative, and versatile benchmark for assessing user profiling approaches.

\subsection{Prompt Template}

\begin{figure*}[ht]
  \centering
  \includegraphics[width=\linewidth]{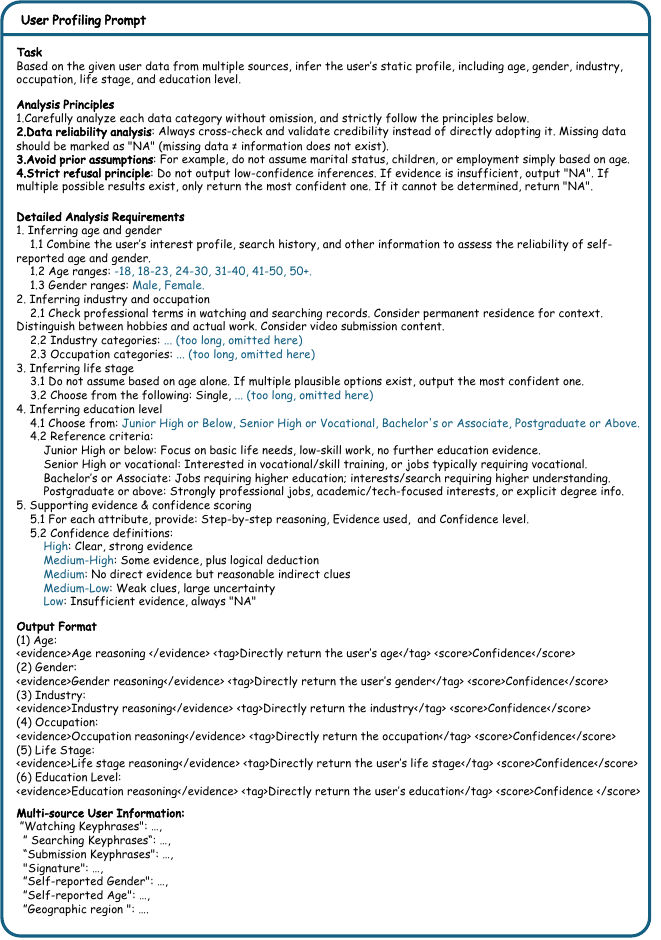}
    \caption{Prompt template of \sysname.}
    \label{fig: prompt appendix}
\end{figure*}

Figure~\ref{fig: prompt appendix} illustrates the prompt template used in our experiments. 
The template mainly consists of task description, analysis principles, profiling system, confidence scoring criteria, output format, and user input. 
Among them, the analysis principles emphasize rejection rules and guidance for reliable responses. 
This design aligns with the selective classification setting, where the model is encouraged to output 'NA' when sufficient reasoning evidence is lacking.
Due to space limitations, some attributes are omitted from the figure; however, the complete prompt includes the full profiling taxonomy.

\clearpage

\begin{figure*}[t]
  \centering
  \includegraphics[width=\linewidth]{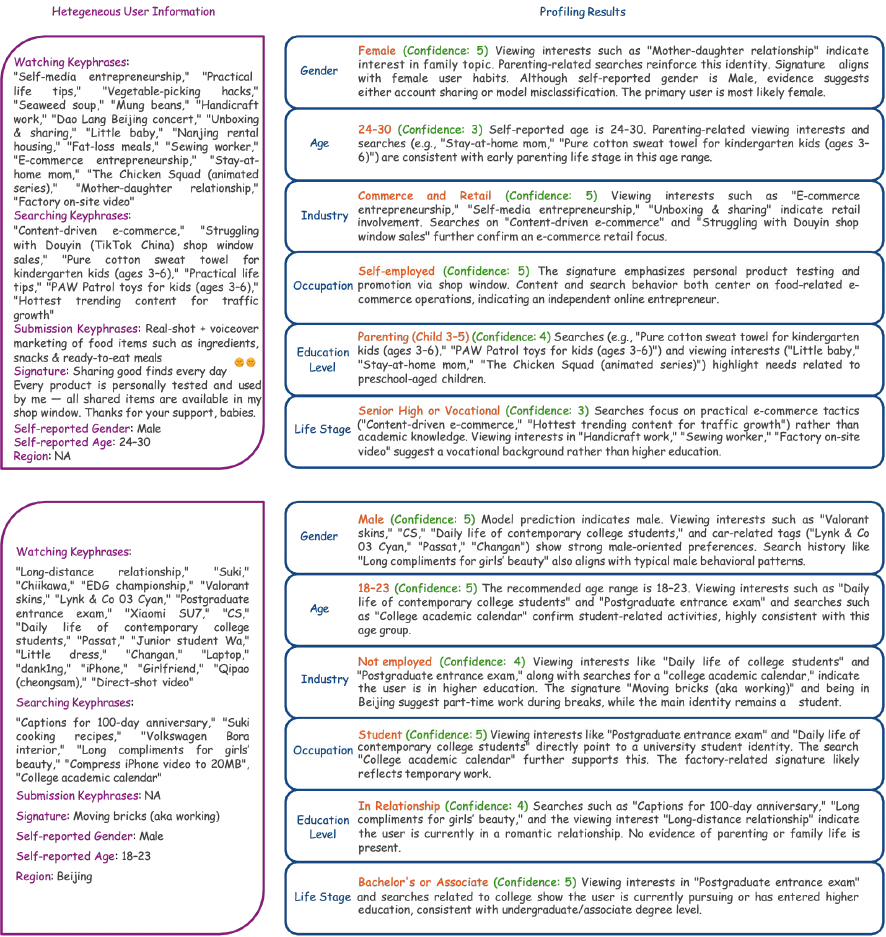}
    \caption{User Profiling Demos.}
    \label{fig: demo appendix}
\end{figure*}

\FloatBarrier

\begin{figure*}[t]
  \centering
    \begin{subfigure}{.49\textwidth}
    \includegraphics[width=\linewidth]{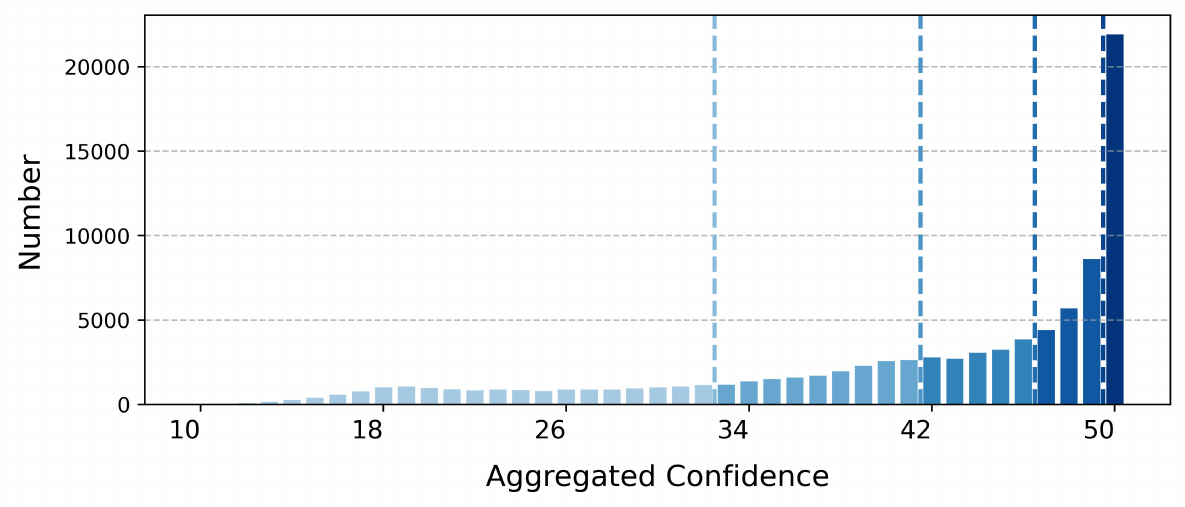}
    \caption{Confidence calibration for gender}
    \end{subfigure}
    \hfill 
    \begin{subfigure}{.49\textwidth}
    \includegraphics[width=\linewidth]{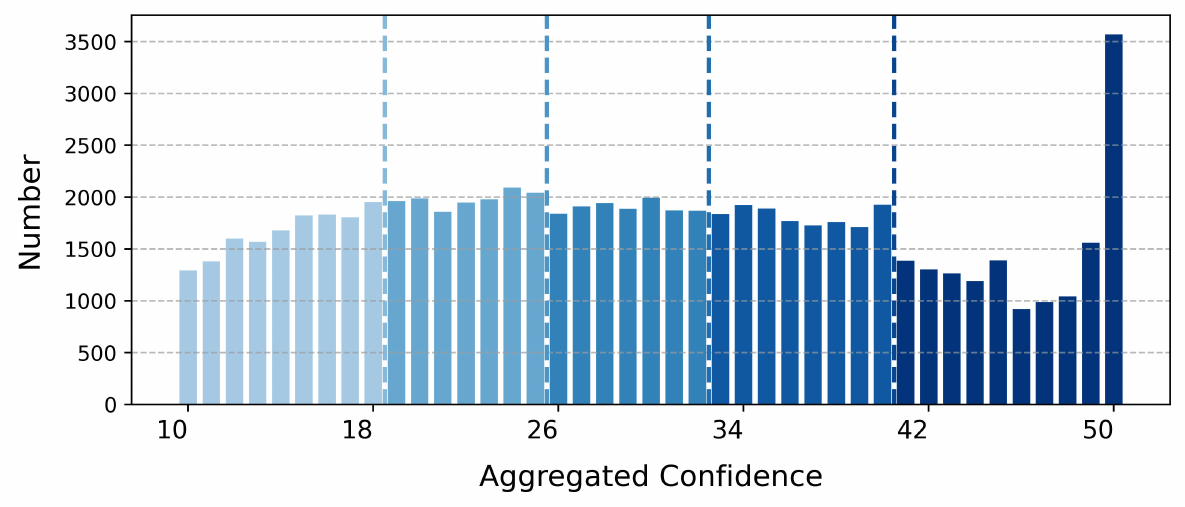}
    \caption{Confidence calibration for industry}
    \end{subfigure}
    \hfill 
    \begin{subfigure}{.49\textwidth}
    \includegraphics[width=\linewidth]{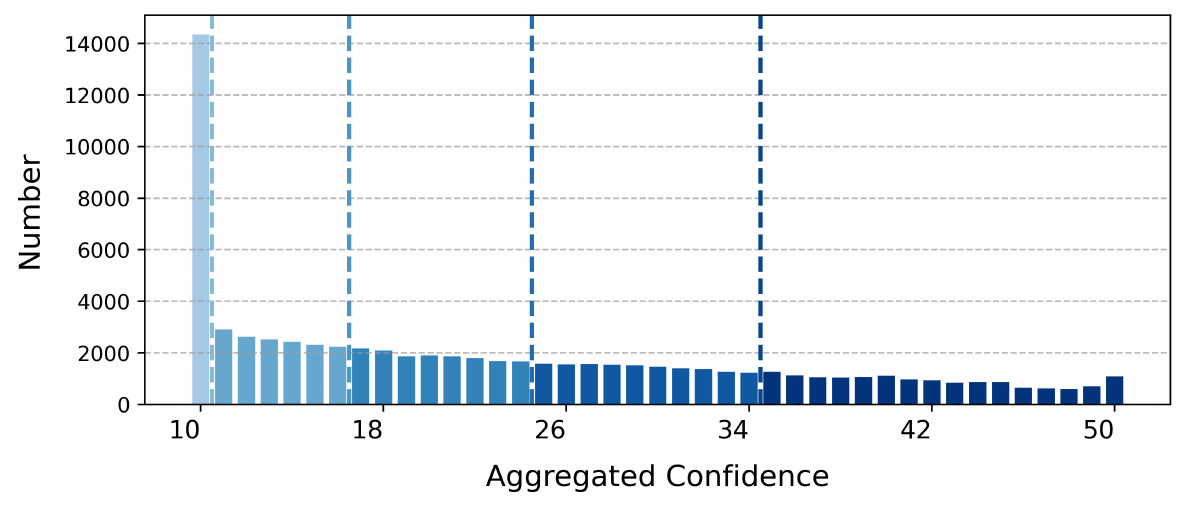}
    \caption{Confidence calibration for occupation}
    \end{subfigure}
    \hfill 
    \begin{subfigure}{.49\textwidth}
    \includegraphics[width=\linewidth]{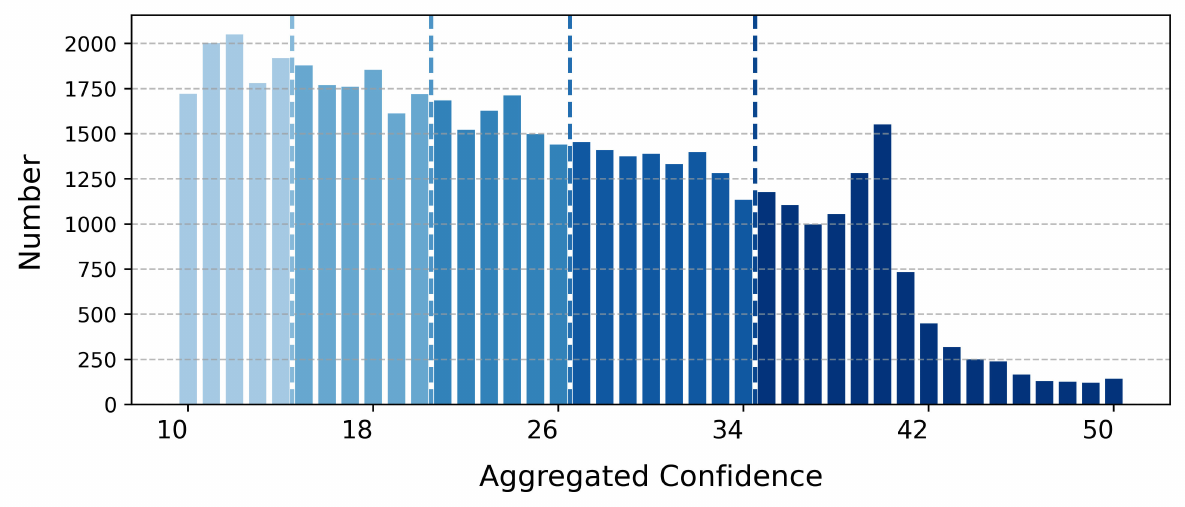}
    \caption{Confidence calibration for education level}
    \end{subfigure}
  \caption{
  Confidence distribution and calibration.
  }
  \label{fig: confidence calibration appendix}
\end{figure*}

\subsection{User Profiling Demos}

Figure~\ref{fig: demo appendix} presents two real cases of profiling inference generated by \sysname. 
Given multi-source user information, the system conducts reasoning and analysis across six attribute dimensions, providing a holistic user profile. 
In the first case, the user is identified as a female who both operates a shop on the platform and is engaged in childcare. 
Interestingly, although the self-reported gender is marked as male, \sysname~ is able to infer the correct gender by leveraging contextual evidence such as interest in female-oriented content, parenting tips, and mother–daughter relationship discussions. 
This demonstrates the model’s ability to resolve inconsistencies and prioritize reasoning over noisy self-reports.  
In the second case, the user is inferred to be a male university student in a romantic relationship. 
The inference is supported by frequent interests in topics such as dating, automobiles, gaming, and college courses. 
Moreover, although the user’s personal signature mentions "moving bricks", the model reasonably interprets this as an indicator of an internship or part-time work, with the primary life stage still classified as a student. 
Overall, these demos highlight the interpretability and robustness of \sysname~ in handling real-world, noisy, and multi-faceted user information.

\subsection{Confidence Calibration}

Figure \ref{fig: confidence calibration} presents the calibration results for age and life stage. 
For the remaining attributes, including gender, industry, occupation, and education level, the corresponding distributions are shown in Figure \ref{fig: confidence calibration appendix}. 
As can be seen, the overall confidence distribution is highly uneven, which necessitates a calibration process. 
Specifically, the raw confidence scores are partitioned into five intervals with approximately equal sample sizes, enabling a balanced representation across different confidence levels.

\subsection{Detailed Experimental Results}

We report the precision, recall, and F1 scores of \sysname~(after SFT and after SFT+RL) under varying confidence thresholds $\tau$. 
On the one hand, by incorporating confidence and thresholding, we enable flexible precision–recall trade-offs for each attribute. 
As expected, higher thresholds yield higher precision but lower recall, while lower thresholds achieve the opposite, providing practitioners with a controllable balance between reliability and coverage. 
Moreover, since the confidence values have been calibrated in advance, the resulting distribution across thresholds is more balanced for each tags, which ensures that evaluation results are not biased toward over-confident predictions. 
On the other hand, the results demonstrate that reinforcement learning consistently improves performance across all attributes and confidence levels, highlighting both the effectiveness and robustness of our approach.

\begin{table*}[ht]
\centering
\setlength{\tabcolsep}{8pt}
\resizebox{\textwidth}{!}{
\begin{tabular}{cc|cccccc|c}
\toprule
\multicolumn{1}{c}{\textbf{$\tau$}} & \multicolumn{1}{c|}{\textbf{Metrics}} & \textbf{Age} & \textbf{Gender} & \textbf{Industry} & \textbf{Occupation} & \textbf{Education Level} & \textbf{Life Stage} & \textbf{Average} \\
\midrule
\multicolumn{9}{c}{\cellcolor{yellow!10}{\sysname~{(\textit{w.} SFT)}}} \\
\midrule
\multirow{3}{*}{1}  
    & Precision & 72.14 & 92.84 & 56.27 & 54.01 & 47.12 & 48.18 & 61.76 \\
    & Recall    & 99.93 & 99.95 & 85.96 & 75.68 & 67.63 & 54.97 & 80.69 \\
    & F1 Score  & 83.79 & 96.26 & 68.02 & 63.03 & 55.54 & 51.35 & 69.67 \\
\midrule
\multirow{3}{*}{2}  
    & Precision & 75.52 & 94.60 & 59.47 & 58.87 & 47.94 & 49.47 & 64.31 \\
    & Recall    & 89.90 & 90.87 & 73.82 & 61.77 & 56.73 & 44.62 & 69.62 \\
    & F1 Score  & 82.08 & 92.70 & 65.87 & 60.29 & 51.97 & 46.92 & 66.64 \\
\midrule
\multirow{3}{*}{3}  
    & Precision & 76.77 & 95.91 & 64.53 & 63.68 & 47.95 & 49.10 & 66.32 \\
    & Recall    & 78.37 & 83.12 & 59.12 & 50.99 & 45.17 & 35.59 & 58.73 \\
    & F1 Score  & 77.56 & 89.06 & 61.71 & 56.63 & 46.52 & 41.27 & 62.12 \\
\midrule
\multirow{3}{*}{4}  
    & Precision & 77.49 & 97.03 & 69.40 & 67.67 & 50.59 & 48.36 & 68.42 \\
    & Recall    & 62.53 & 72.98 & 46.87 & 41.52 & 28.88 & 23.15 & 45.99 \\
    & F1 Score  & 69.21 & 83.30 & 55.95 & 51.46 & 36.77 & 31.31 & 54.67 \\
\midrule
\multirow{3}{*}{5}  
    & Precision & 78.95 & 97.48 & 78.46 & 79.08 & 61.56 & 48.96 & 74.08 \\
    & Recall    & 46.81 & 66.56 & 28.38 & 22.80 & 13.50 & 10.46 & 31.42 \\
    & F1 Score  & 58.77 & 79.11 & 41.68 & 35.40 & 22.14 & 17.24 & 42.39 \\
\midrule
\multicolumn{9}{c}{\cellcolor{yellow!10}{\sysname~{(\textit{w.} SFT+RL)}}} \\
\midrule
\multirow{3}{*}{1}  
    & Precision & 73.29 & 92.96 & 61.24 & 57.68 & 49.20 & 54.59 & 64.83 \\ 
    & Recall    & 99.87 & 99.87 & 86.25 & 87.99 & 72.28 & 60.79 & 84.51 \\ 
    & F1 Score  & 84.54 & 96.29 & 71.63 & 69.68 & 58.55 & 57.52 & 73.03 \\ 
\midrule
\multirow{3}{*}{2}  
    & Precision & 75.11 & 94.18 & 61.48 & 61.17 & 50.92 & 55.17 & 66.34 \\ 
    & Recall    & 93.43 & 93.50 & 78.15 & 74.63 & 55.30 & 55.36 & 75.06 \\ 
    & F1 Score  & 83.27 & 93.84 & 68.82 & 67.23 & 53.02 & 55.27 & 70.24 \\ 
\midrule
\multirow{3}{*}{3}  
    & Precision & 76.48 & 96.15 & 67.15 & 65.17 & 52.84 & 58.84 & 69.44 \\ 
    & Recall    & 84.36 & 82.76 & 60.35 & 56.91 & 40.93 & 45.67 & 61.83 \\ 
    & F1 Score  & 80.23 & 88.96 & 63.57 & 60.76 & 46.13 & 51.42 & 65.18 \\ 
\midrule
\multirow{3}{*}{4}  
    & Precision & 77.74 & 97.76 & 71.72 & 69.82 & 56.05 & 58.96 & 72.01 \\ 
    & Recall    & 69.65 & 71.09 & 42.98 & 41.29 & 22.79 & 25.40 & 45.53 \\ 
    & F1 Score  & 73.47 & 82.32 & 53.75 & 51.89 & 32.40 & 35.51 & 54.89 \\ 
\midrule
\multirow{3}{*}{5}  
    & Precision & 77.27 & 98.21 & 81.07 & 81.57 & 72.99 & 51.74 & 77.14 \\ 
    & Recall    & 52.74 & 65.43 & 23.49 & 18.70 & 13.51 & 8.22 & 30.35 \\ 
    & F1 Score  & 62.69 & 78.54 & 36.43 & 30.43 & 22.80 & 14.19 & 40.84 \\
\bottomrule
\end{tabular}
}
\caption{
    The detailed performance of \sysname.
}
\label{tab:detailed_perf2}
\end{table*}

\end{document}